\documentclass[letterpaper, 10 pt, conference]{ieeeconf}  
\usepackage{tikz}
\usepackage{color}
\usepackage{amsmath}
\usepackage[numbers]{natbib}
\usepackage{hyperref}
\usepackage{float}
\usepackage[linesnumbered,ruled,vlined]{algorithm2e}
\usepackage[pages=some]{background} 

\usetikzlibrary{math, shapes, arrows, calc, angles, quotes, arrows.meta, fit, positioning, shapes.geometric}

\IEEEoverridecommandlockouts                              

\title{\LARGE \bf
Speeding Up Path Planning via Reinforcement Learning in MCTS for Automated Parking 
}

\author{Xinlong Zheng$^{*}$, Xiaozhou Zhang$^{*}$, Donghao Xu \\
DeepRoute.Ai \\
3125 Skyway Ct, Fremont, CA 94539, USA \\
{\tt\small \{xinlongzheng, xzzhang\}@alumni.upenn.edu}, {\tt\small donghaoxu@deeproute.ai}
\thanks{*Equal Contribution}
}

\begin{document}

\maketitle

\thispagestyle{empty}
\pagestyle{empty}

\begin{abstract}

In this paper, we address a method that integrates reinforcement learning into the Monte Carlo tree search to boost online path planning under fully observable environments for automated parking tasks. Sampling-based planning methods under high-dimensional space 
can be computationally expensive and time-consuming. State evaluation methods are
useful by leveraging the prior knowledge into the search steps, making the process faster in a real-time system.
Given the fact that automated parking tasks are often executed under complex environments,  a solid but lightweight heuristic guidance is challenging to compose in a traditional analytical way. To overcome this limitation, we propose a reinforcement learning pipeline with a Monte Carlo tree search under the path planning framework. By iteratively learning the value of a state and the best action 
among samples from its previous cycle's outcomes, we are able to model a value estimator and a policy generator for given states. By doing that, we build up a balancing mechanism between exploration and exploitation, speeding up the path planning process while maintaining its quality without using human expert driver data.

\end{abstract}

\section{INTRODUCTION}
\backgroundsetup{opacity=1, scale=1, angle=0, contents={%
\color{black}
\fboxsep=0mm
\fboxrule=0mm
\begin{tikzpicture}[remember picture,overlay]\node[anchor=south, yshift=20pt] at (current page.south) {\fbox{\parbox{\dimexpr\textwidth-\fboxsep-\fboxrule\relax}{\small{DOI: 10.1109/IROS58592.2024.10802417}}}};\end{tikzpicture}
\begin{tikzpicture}[remember picture,overlay]\node[anchor=north,yshift=-30pt] at (current page.north) {\fbox{\parbox{\dimexpr\textwidth-\fboxsep-\fboxrule\relax}{\small{2024 IEEE/RSJ International Conference on Intelligent Robots and Systems (IROS)}}}};\end{tikzpicture}
}}
\BgThispage

In general, an automated parking task is primarily separated into sub-modules such as perception, localization, planning, control, etc. 
The appearance of novel perception frameworks such as Bird's-Eye-View\cite{philion2020lift} and occupancy networks\cite{wei2023surroundocc} brings clarity into sensing the environments in an 
online automated parking task. There are also well-studied methods of localization to achieve a high-precision result. The performance of an automated parking execution heavily relies on the outcomes from the planning and control module.

Given a fully observable environment, the goal of 
the planning module is to generate an optimal trajectory for autonomous vehicles. The trajectory consists of several 
points in the spatio-temporal space, combining both path and speed information. Common methodologies applied on a real-time planner partition 
the spatio-temporal domain into path planning and speed planning in order to reduce the complexity and computational load. 
The speed planning under parking scenarios can be relatively easy compared to on-road driving scenarios since the planning time horizon is limited, the search space is smaller, 
and there are fewer interactive objects involved.

\begin{center}
   \begin{figure}[thpb]
      \centering
      \begin{tikzpicture}

         \draw (0, 0) node {\includegraphics[width=2cm]{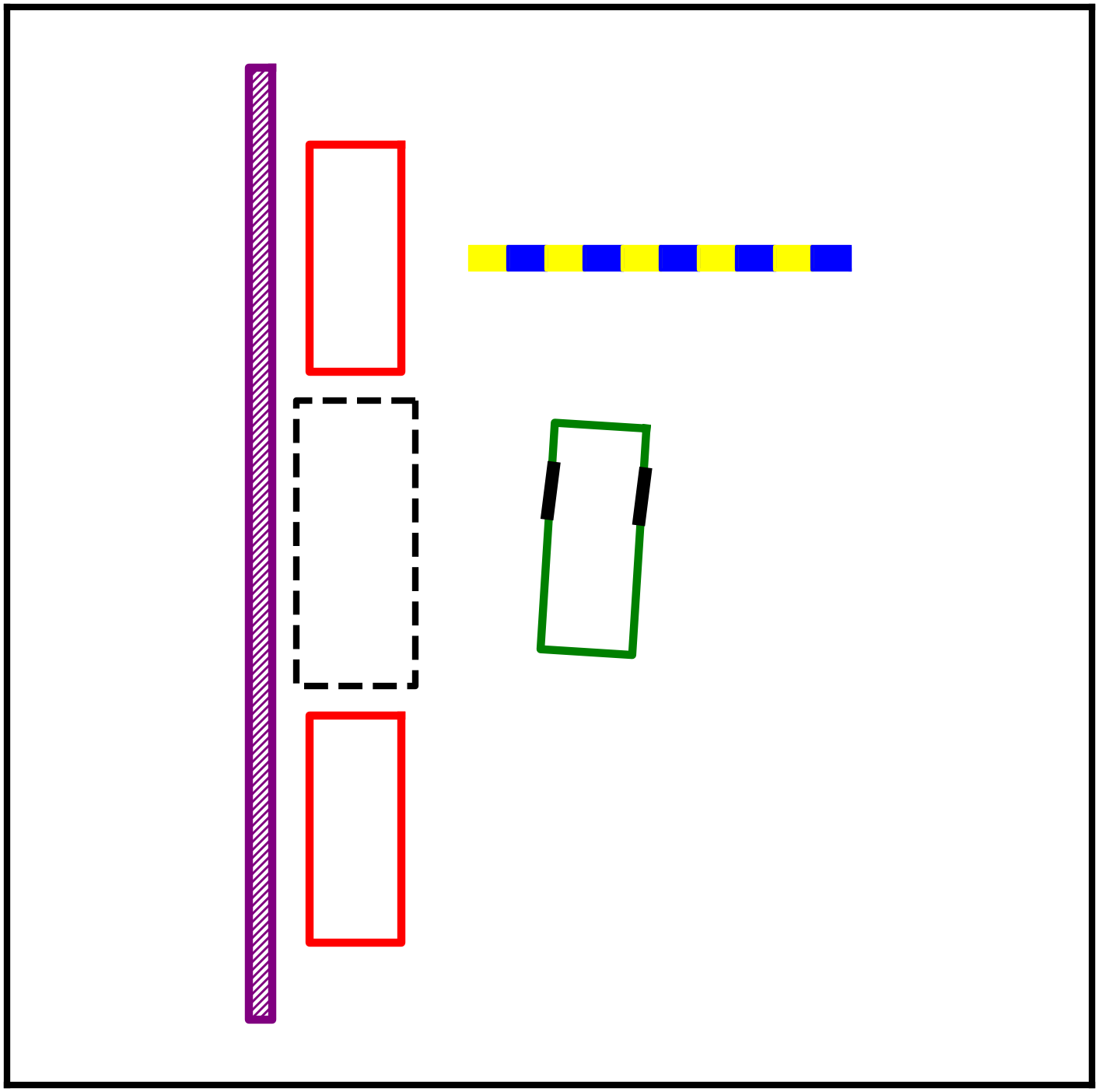}} node[above=1cm] {$s_0$};
         \draw (5, 0) node {\includegraphics[width=2cm]{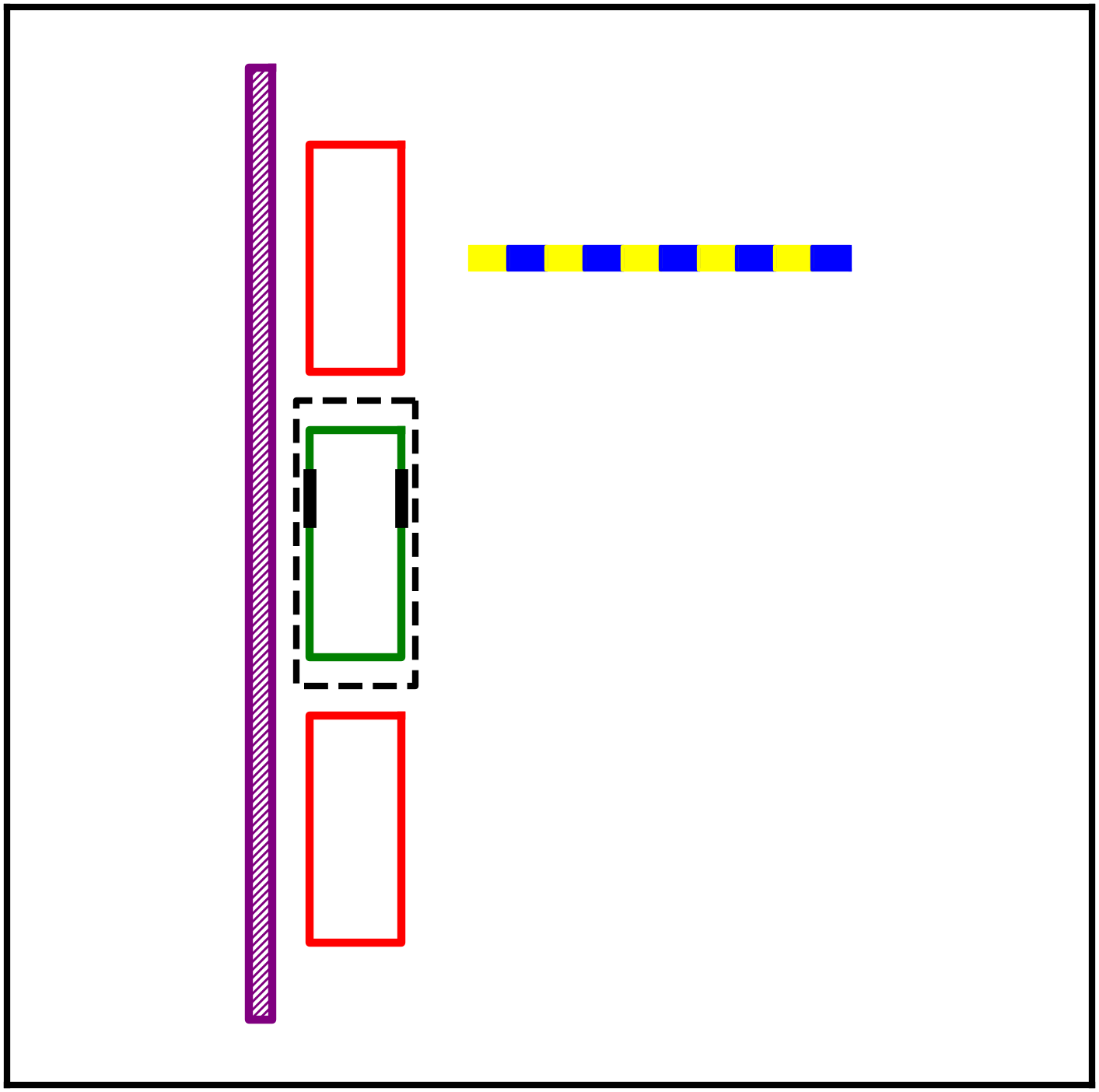}} node[above=1cm] {$s_\text{T}$};
         \draw (2.5, 0) node {\textbf{...}} node[above=1mm] {$a_t\sim\textbf{PUCT}_t$};
         \begin{scope}[>=Latex]
           \draw[->, thick] (1.1, 0.0) --  (2.2, 0);
           \draw[->, thick] (2.8, 0.0) --  (3.9, 0);
           \draw[->, ultra thick] (5, -1.1) --  (5, -4.4);
         \end{scope}
       
         \draw[fill=blue!10, thick, rounded corners=5pt] (1.2, -1.2)  rectangle (3.8, -4.2);
       
         \node[draw, circle, minimum size=0.2cm, inner sep=0pt, fill=black!30, thick] (a) at (2.5, -2.5) {};
         \node[draw=black!30, circle, minimum size=0.2cm, inner sep=0pt, fill=white] (b) at (1.75, -3) {};
         \node[text=black!30] (c) at (2.25, -3) {...};
         \node[draw, circle, minimum size=0.2cm, inner sep=0pt,  fill=black!30, thick] (d) at (2.75, -3) {};
         \node[draw=black!30, circle, minimum size=0.2cm, inner sep=0pt, fill=white] (e) at (3.25, -3) {};
         \node[text=black!30] (f) at (1.75, -3.5) {...};
         \node[draw, circle, minimum size=0.2cm, inner sep=0pt,  fill=black!30, thick] (g) at (2.5, -3.5) {};
         \node[draw=black!30, circle, minimum size=0.2cm, inner sep=0pt, fill=white] (h) at (3, -3.5) {};
         \node[draw, circle, minimum size=0.2cm, inner sep=0pt,  fill=black!30, thick] (i) at (2.5, -4) {};
       
         \begin{scope}[>=Stealth]
         \draw[->, color=black!30] (a)--(b);
         \draw[->, color=black!30] (b)--(f);
         \draw[->, thick] (d)--(a);
         \draw[->, color=black!30] (a)--(e);
         \draw[->, color=black!30] (d)--(h);
         \draw[->, thick] (i)--(g);
         \draw[->, thick] (g)--(d);
         \end{scope}
         \node at (2, -1.8) {\tiny\textbf{MCTS}};
       
         \draw[-Stealth, color=red!30, ultra thick] (2.5, -1) -- (2.5, -0.2);
       
         \draw (5, -5.5) node (fff) {\includegraphics[width=2cm]{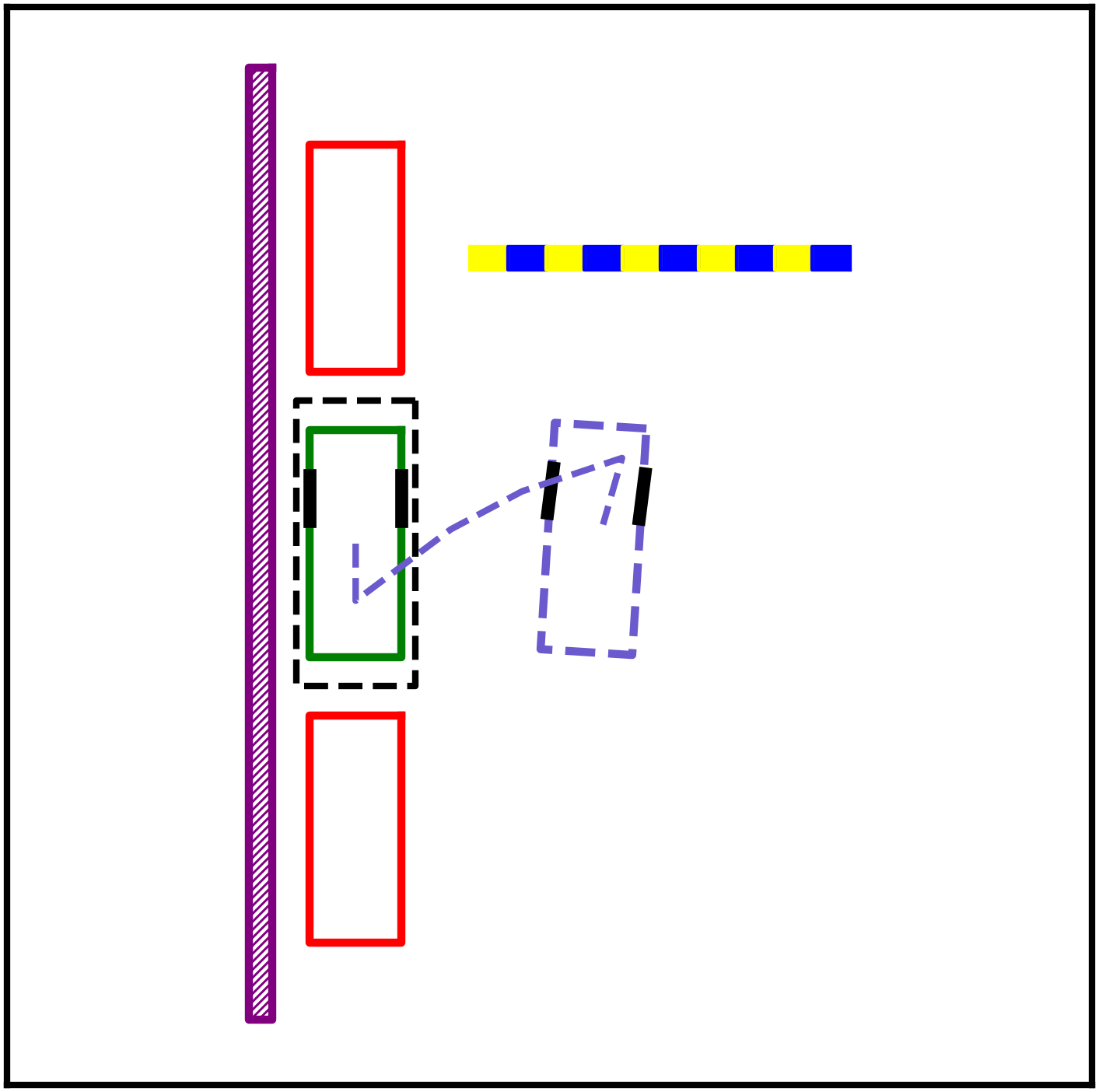}} node[below=1cm] {$z$};
       
         \draw (0, -7.2) node (ff) {\includegraphics[width=1cm]{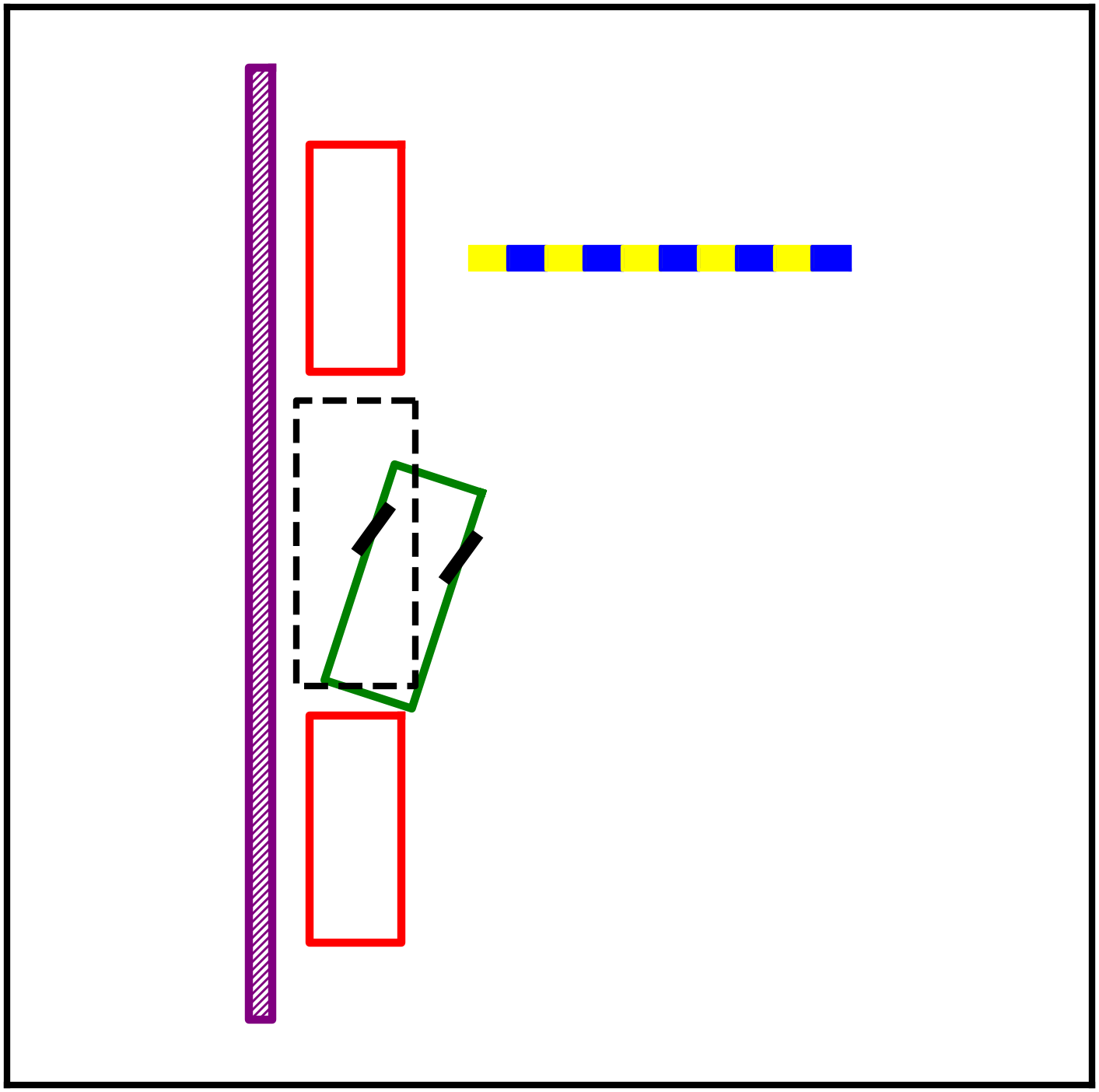}} node[below=0.5cm] {$s_t$};
       
         \node[trapezium, fill=black!25, rotate=-90, minimum width=40, outer sep=2pt] (t) at (0, -5.5) {};
       
         \node[thick] at (-0.6, -5.5) {\tiny {${\boldsymbol{f_\theta}}$}};
       
         \draw[-Stealth, thick] (ff) -- (t);
       
         \node[draw, circle, inner sep=0pt, minimum size=0.1cm, fill=yellow!30] (pt) at (1, -5) {\tiny $\boldsymbol{p_t}$};
         \node[draw, circle, inner sep=0pt, minimum size=0.1cm, fill=yellow!30] (vt) at (1, -6) {\tiny $v_t$};
       
         \node (app1) at (1.3, -5) {\tiny $\approx$};
         \node (app2) at (1.3, -6) {\tiny $\approx$};
         
         \node[draw, circle, inner sep=0pt, minimum size=0.1cm, fill=green!30] (pit) at (1.6, -5) {\tiny $\boldsymbol{\pi_t}$};
         \node[draw, circle, inner sep=0pt, minimum size=0.1cm, fill=green!30] (rt) at (1.6, -6) {\tiny $r_t$};
       
         \draw[dashed, ->] (t) -- (pt);
         \draw[dashed, ->] (t) -- (vt);
       
         \draw[dashed, ->] (fff) -- (pit);
         \draw[dashed, ->] (fff) -- (rt);
         \draw[dashed, ->] (fff) -- (ff);
       
         \draw[-Stealth, ultra thick, color=blue!60] (t) .. controls (0.0, -2.7) .. (1.1, -2.7);
         
       
         \draw[dashed, thick] (-1.5,1.5) rectangle (6.5, -4.3);
         \draw[dashed, thick] (-1.5,-4.3) rectangle (6.5, -8.2);
       
         \node at (-1.2, -4.1) {\tiny \textbf{(a)}};
         \node at (-1.2, -4.5) {\tiny \textbf{(b)}};
       
       
       \end{tikzpicture}
 \caption{Reinforcement learning integrated MCTS in path planning tasks. (a) The agent plans its move under the guidance of MCTS. $s_0$ is the start state and $s_\text{T}$ is the destination state. $a_t$ is the action takes at time $t$ selected by PUCT\cite{rosin2011multi}. (b) Neural network training against the previously produced results. $z$ is the terminated tree that can be used to generate training input state $s_t$, training label $\boldsymbol{\pi}_t$, and $r_t$. $f_\theta$ is the neural network projecting $s_t$ to policy distribution $\boldsymbol{p_t}$ and value $v_t$.}
 \label{figurelabel0}
\end{figure}
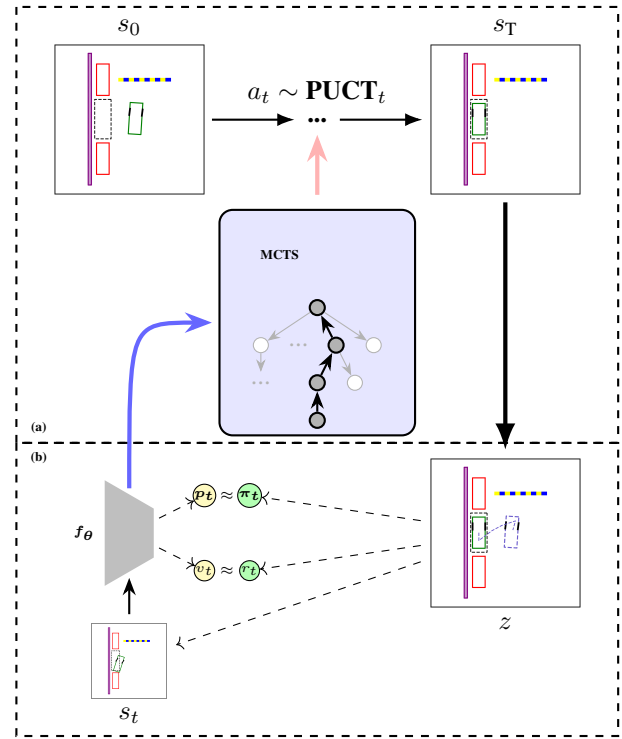
\end{center}

Sampling-based approaches are the most common method employed on the path planning task 
in parking scenarios, and it can be followed by a numeric optimization after. The performance of 
such an algorithm relies heavily on its sampling density --- the denser the sampling, the algorithm is more likely to converge towards a global optimal solution. Under the parking tasks, especially those complex ones,
 the vehicle would like to use most of the spare space with bi-directional maneuvers to reach the destination spot with minimal effort while ensuring a comfortable experience for clients. In order to get a satisfying result, a higher dimension of sampling space and a compact sampling strategy are needed, which makes the planning process time-consuming.

Many methods have been introduced to reduce the time consumption for such algorithms in a real-time system. Most of them aim to enhance the exploitation capability of 
the algorithm, by either trimming out infeasible samples or prioritizing the promising ones with a heuristic estimate. However, too much preference on the exploitation 
side would decrease the ability to explore and lead to sub-optimal solutions.

Monte Carlo tree search (MCTS)\cite{kocsis2006bandit}, as an orderly sampling-based method, shows its strength in solving long-horizon planning tasks like path planning. 
However, without the significant guidance of prior domain knowledge, the iterative tryouts from its cycled operations make the planning time too long to run in the real-time system. 
In our approach, we would like to leverage the capability of reinforcement learning to provide domain knowledge guidance for MCTS, boosting up its planning time and finding the equilibrium of searching process, without the help from human driver data, as what is shown in Fig. 1.

\section{RELATED WORKS}
The path planning problem in automated parking tasks has been studied for decades. It can be easily treated as a Euclidean geometry problem with constraints. By putting pieces of straight lines and basic curves together, one can construct a path by connecting a series of pre-designed points. The curves are mainly circle arcs that can be computed to meet the basic kinematic constraints like maximum curvature. \cite{sungwoo2011easy} \cite{paromtchik1996autonomous} have implemented these geometric methods in different types of parking scenarios. The computing time of the geometric method is very short, but it only works well for easy parking tasks, since there is not enough variation in the shape of the path.

The sampling-based approach is well-studied as well. In \cite{pepy2006path}, implementation of one of the most popular random-based sampling algorithms is given: the Rapidly-exploring Random Tree (RRT)\cite{lavalle2001rapidly} for planning on dynamic vehicle model, and it has been used by \cite{kuwata2009real} in automated parking situation. RRT has a simple logic and is easy to implement, but its randomness leads to a lack of stability and is not guaranteed to find an optimal solution.

Besides the random one, another popular orderly-based sampling algorithm is A*\cite{doran1966experiments} and its variant.
\cite{dolgov2008practical} comes up with a Hybrid-State A* Search which associates a continuous state with each sample, and uses it to solve a perpendicular parking task. The process of A* is expedited by the prior knowledge known as heuristic function\cite{doran1966experiments}. In terms of automated parking, the study on its prior knowledge is mainly focused on the shape properties of the curves, such as the Reeds-Shepp (RS) curve\cite{reeds1990optimal}, Bezier curve\cite{liang2012automatic}, etc. However, the curve's shape property becomes less useful when there are lots of obstacles in the parking environment.

Another way to formulate the corresponding prior knowledge considering both the curve's shape and obstacles is through supervised learning from human expert driver data like \cite{notomista2017machine}. To achieve a well-rounded performance, a large amount of high-quality human data is required and these are often expensive. 
Prior knowledge can also be enhanced using reinforcement learning strategies. Since the advent of AlphaGo \cite{silver2017mastering}, researchers have been working to solve long-horizon planning tasks in the field of autonomous driving, such as those discussed in \cite{weingertner2020monte}. \cite{song2022time} managed to address autonomous parking tasks with MCTS, but their solution was limited to parallel parking scenarios.

\section{PRELIMINARIES}
\subsection{Problem Formulation}
Assume that the automated parking task is executed at a low constant speed in an observable and consistent environment. 
The path planning in such task can be considered as a Markov Decision Process (MDP) with 
certain $\mathcal{S}$, $\mathcal{A}$, $\mathcal{T}$,  $\mathcal{O}$, and $\mathcal{R}$. The goal of this process for autonomous vehicles is 
to reach the destination pose from a starting pose while minimizing the corresponding cost of doing that. The MDP can be formulated as follows.

Let $\mathcal{S}$ be the state space of the agent (autonomous vehicle). A state $s_t$ can be defined as the pose of the vehicle as $(x_t, y_t, \phi_t)$ at time $t$, where $x$ and $y$ are the coordinates 
of the rear axis center in a 2D Cartesian space and $\phi$ is the heading of the vehicle. 
\begin{center}
   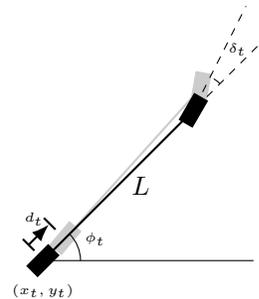
\begin{figure}[thpb]
      \centering
   \begin{tikzpicture}
      \filldraw[fill=black!20, black!20, rotate around={50:(0.5, 0.4)}] (0.3,0.3) rectangle (0.7,0.5);
      \filldraw[fill=black!20, black!20, rotate around={80:(2.3, 2.4)}] (2.1,2.3) rectangle (2.5,2.5);
      \draw[black!20, thick] (2.3,2.4)--(0.5,0.4);
      \begin{scope}[>=Latex]
      \draw[|->|, thick] (0.0, 0.3) -- (0.3, 0.6);
      \draw (0.08, 0.65) node {\tiny $d_t$};
      \end{scope}
    
      \filldraw[fill=black, rotate around={45:(0.2, 0.1)}] (0,0) rectangle (0.4,0.2);
      \filldraw[fill=black, rotate around={60:(2.2, 2.1)}] (2,2) rectangle (2.4,2.2);
      \draw[black, thick] (2.2,2.1)--(0.2,0.1);
      \draw (1.5,1.1) node {$L$};
      \draw (0.2, -0.3) node {\tiny $({x_t, y_t})$};
      \draw[dashed] (3.2, 3.1) coordinate (A) -- (2.2, 2.1) coordinate (B) -- (2.9, 3.485) coordinate (C) pic ["\tiny $\delta_{t}$", draw, angle eccentricity=2] {angle};
      \draw[] (3, 0.1) coordinate (A) -- (0.2, 0.1) coordinate (B) -- (0.8, 0.7) coordinate (C) pic ["\tiny $\phi_{t}$", draw, angle eccentricity=1.5] {angle};
    \end{tikzpicture}
    \caption{Bicycle vehicle model}
   \label{figurelabel1}
\end{figure}
\end{center}

Let $\mathcal{A}$ be the action space of the agent and $\mathcal{T}$ as the transition function maps a state $s_t$ to $s_{t+1}$ with a given 
action $a_t$. We use the bicycle model as in Fig. 2 for vehicle kinematics under this low-speed parking scenario. By considering the speed as a constant as well, we 
can use the distance gap $d_t$ to denote the gap between different vehicle poses. The motion equations can be written as follows:

\begin{equation} \label{eqno1}
\begin{split}
       & x_{t+1} = x_{t} + d_{t} \cos \phi_{t} \\
       & y_{t+1} = y_{t} + d_{t} \sin \phi_{t} \\
       & \phi_{t+1} = \phi_{t} + \frac{d_t \tan \delta_{t}}{L}
\end{split}
\end{equation}
Equation (1) reveals the composition of action space as an action $a_t = (d_{t}, \delta_t)$ , where $\delta_t$ is the front wheel angle 
of the vehicle. (1) itself is the transition function, where $L$ is the length of the wheel base, with the kinematic and collision constraint taken into account.
In practice, the action space is discretised, which made the state space discrete as well.

Let $\mathcal{O}$ be the observation space. $o_t$ is the observation at time $t$, which is sensed by the perception module formulated as 
the occupancy grids of different layers for interested environment objects of different categories such as parking slots, obstacles, road edges, speed bumps, etc. Since the 
environment is assumed as consistent during the whole process, the $o_{t}$ is a constant at any time $t$ as well.

Let $\mathcal{R}$ be the reward function which maps a corresponding cost by the sub-sequential states the process goes so far and a fixed scalar reward if the agent 
reaches the destination as 

\begin{equation} \label{eqno2}
\begin{split}
     r  & = \mathcal{R}(s_0, s_1, ..., s_t, o) + r_t\\
        & = \mathcal{C}_{\text{safety}}(s_0, s1, ..., s_t, o) \\
        & \ \ \  + \mathcal{C}_{\text{comfort}}(s_0, s1, ..., s_t)\\
        & \ \ \ + \mathcal{C}_{\text{efficiency}}(s_0, s1, ..., s_t) + r_t
\end{split}
\end{equation}
If the agent at $s_t$ reaches the destination pose, $r_t = 1$, otherwise $r_t = 0$. The cost function itself is related to safety, comfort, and efficiency 
of the path formed by traveling through the states. Paths getting closer to the obstacles within a certain threshold would result in more negative safety cost. Paths with redundant wheel or gear moves would result in more negative comfort and efficiency cost .

\subsection{Monte Carlo Tree Search}
Since the discovery of Upper Confidence bounds applied to Trees (UCT)\cite{kocsis2006bandit} was introduced in the year 2006, the Monte Carlo tree search has 
become popular when it comes to finding an optimal solution by searching in high-dimension spaces. Especially after MCTS was used for AI to play long-horizon board games like 
chess and go\cite{silver2016mastering}\cite{silver2017mastering}, it was notably suitable for solving episodic planning tasks like an MDP problem.

The basic idea of MCTS just meets the requirements of the path planning task: it naturally balances between exploitation and exploration by building 
a search tree and iteratively updating its search strategy. The algorithm itself falls into a cycle consisting of the following 
procedures: \textit{selection}, \textit{expansion}, \textit{simulation} and \textit{backpropagation}.

During the \textit{selection} phase, MCTS repeatedly chooses a child node to visit starting from the root node, until it finds a node that is not \textit{expanded}. The \textit{selection} is guided by certain criteria such as the \textit{upper confidence bound} (UCB)\cite{auer2002finite} and the \textit{predictor upper confidence bound} (PUCB)\cite{rosin2011multi}. This criterion primarily aims to strike a balance between exploiting the child with known high rewards, termed exploitation, and exploring lesser-visited children, known as exploration.


If the selected node does not reach the terminal state, it enters the $epansion$ phase and a $simulation$ will be applied to the node. 
A $simulation$ can be carried out with different methods such as random rollout, rule-based rollout, or other estimations to get a reward for the node. 

The estimated reward will be \textit{propagated back} from the \textit{simulated} child all the way back to the root node. Interested properties of the nodes along the retrieved path will be updated respectively as well.


The whole cycle will be repeated until a terminal condition is met. Then the updated preference 
on the nodes from the search tree will be adapted for real-time execution from the 
root node with the same \textit{selection-expansion-simulation-backpropagation} procedures.

\section{METHODOLOGY}

\begin{center}
   \begin{figure}[H]
      \centering
\begin{tikzpicture}[]
   \tikzmath{\scale = 0.8; \xoffset = 7*\scale;\xchild = 0.7*\scale;\ychild=1.6*\scale;\fwidth=1.0*\scale;\yoffset=5.5*\scale;}
   \tikzmath{\crosslength = 0.1*\scale;}
   \tikzmath{\ax=0.0; \ay=-\yoffset; \bx=0.0; \by=0.0; \cx=\xoffset; \cy=0.0; \dx=\xoffset; \dy=-\yoffset;}
   
   \draw (0, 0) node (p01) {\fcolorbox{black}{black}{\includegraphics[width=\fwidth cm]{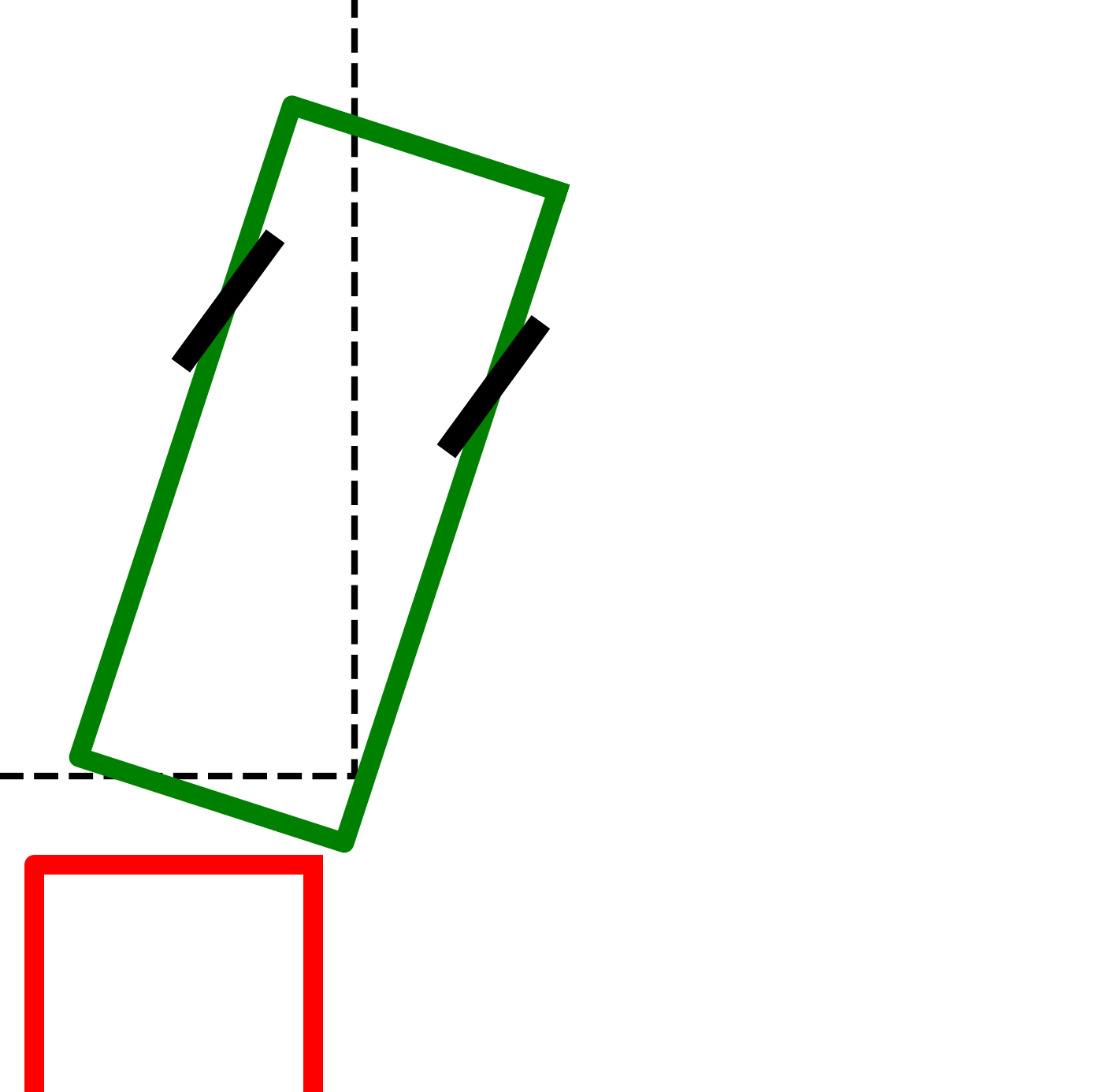}}};
   \draw (\xchild+\bx, -\ychild+\by) node (p11) {\fcolorbox{red!50}{black}{\includegraphics[width=\fwidth cm]{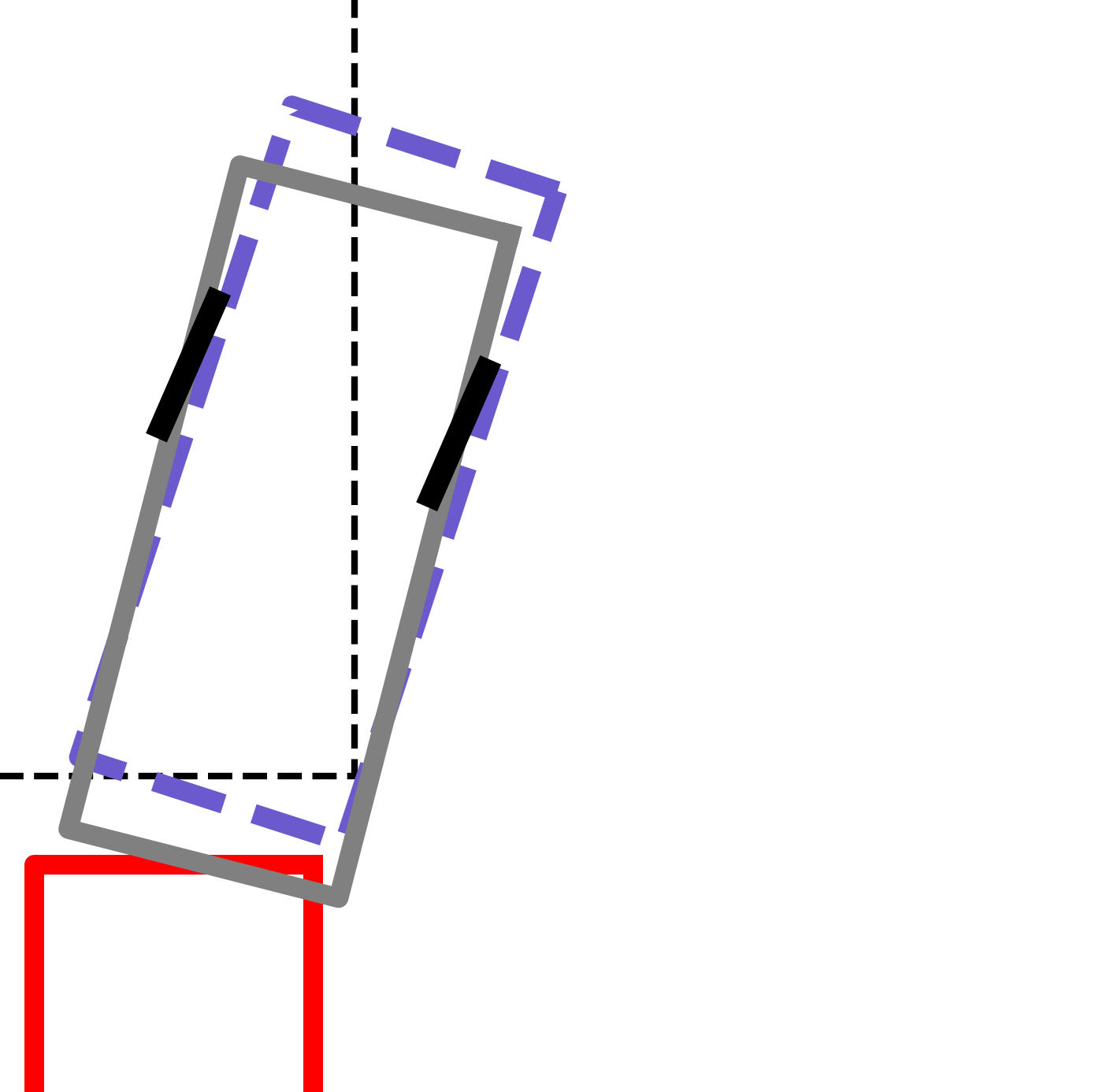}}};
   \draw (-\xchild+\bx, -\ychild+\by) node (p21){\fcolorbox{black}{black}{\includegraphics[width=\fwidth cm]{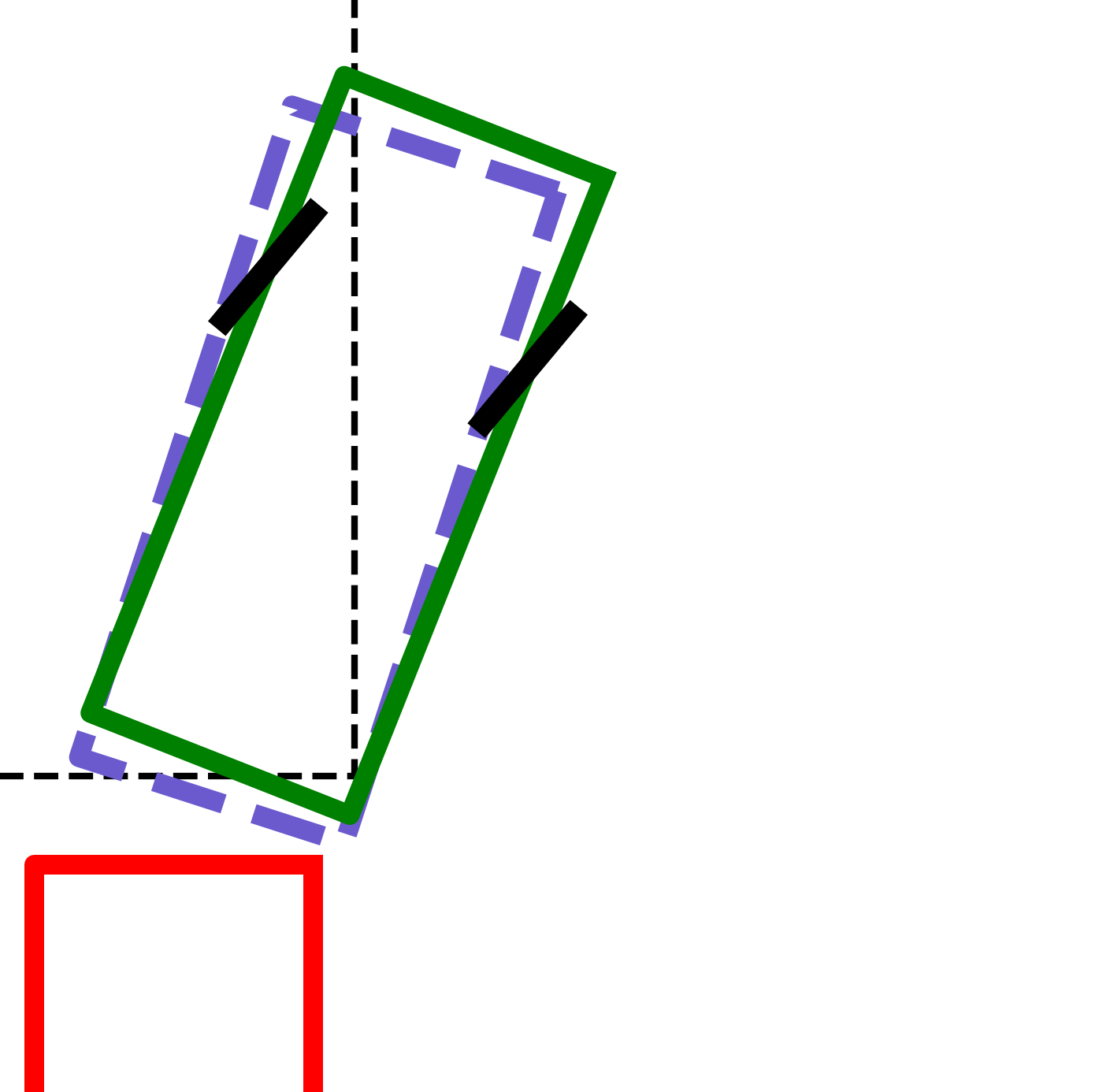}}};

   \draw  (-\xchild-\xchild+\bx, -\ychild-\ychild+\by) node (k01) {\fcolorbox{black!30}{black}{\includegraphics[width=\fwidth cm]{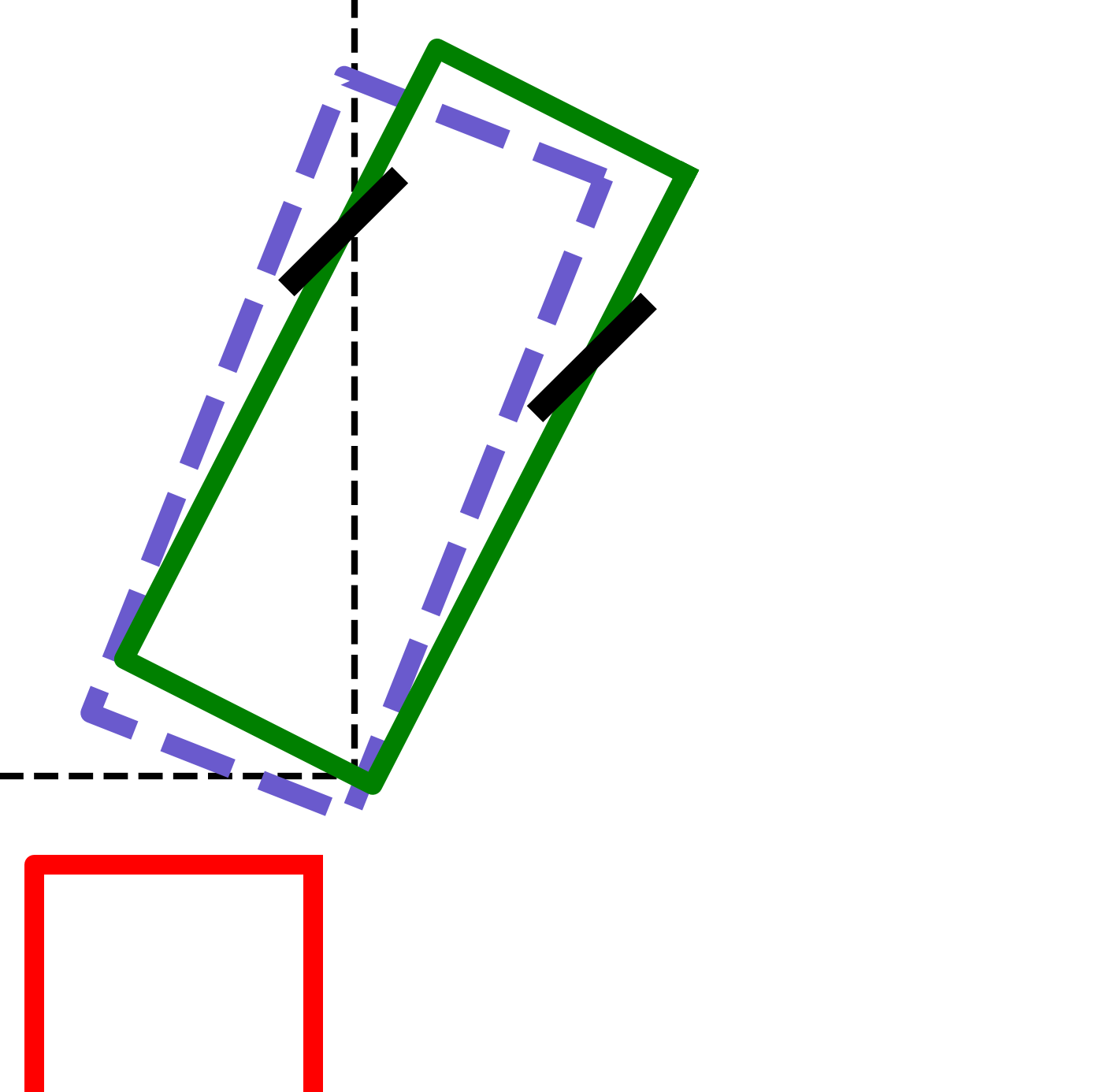}}};
   \draw  (-\xchild+\xchild+\bx, -\ychild-\ychild+\by) node (k11) {\fcolorbox{blue!50}{black}{\includegraphics[width=\fwidth cm]{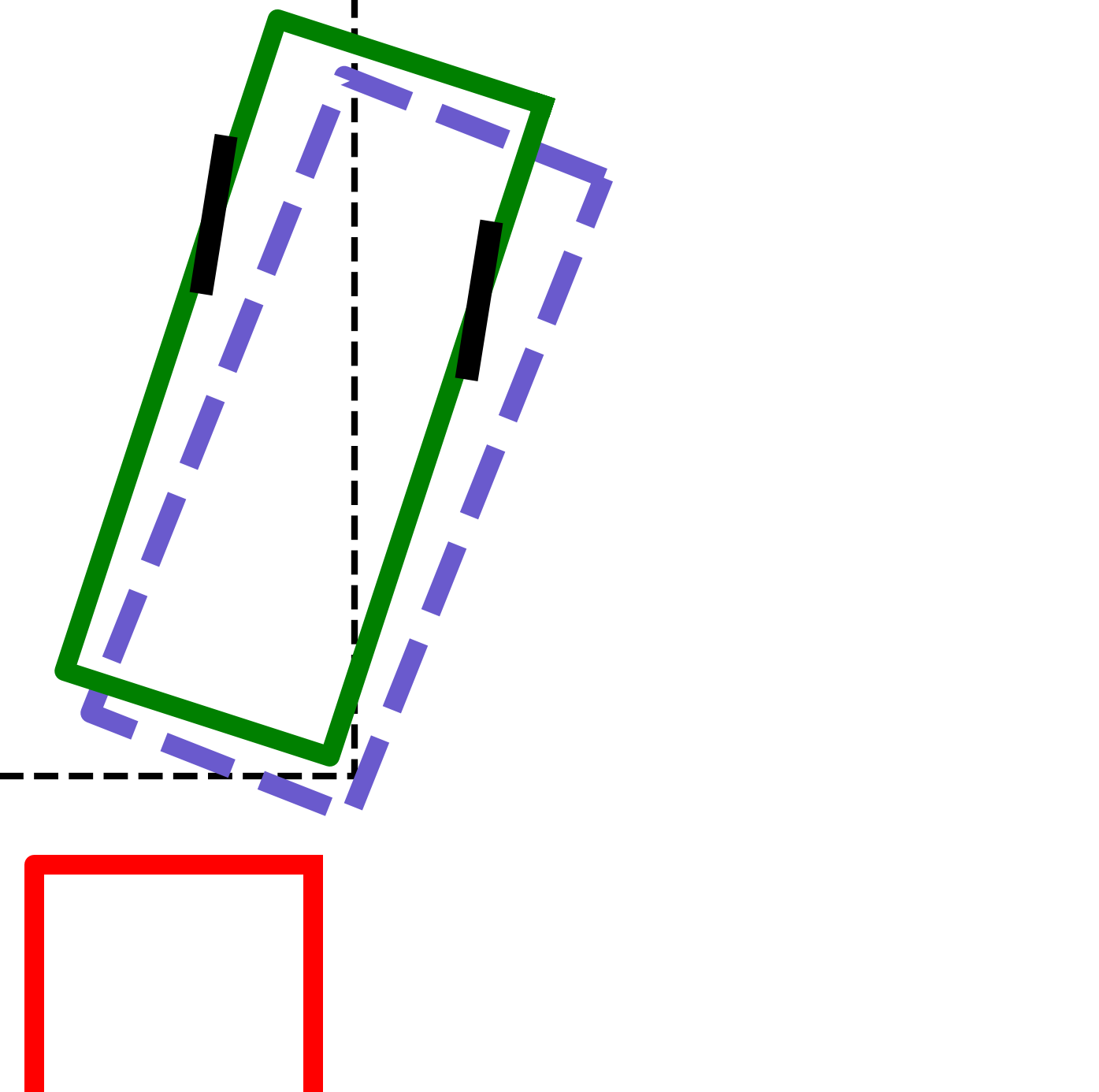}}};
 
   \draw (-\xchild+\bx, -\ychild*3+\by) node (kk01) {\includegraphics[width=\fwidth cm]{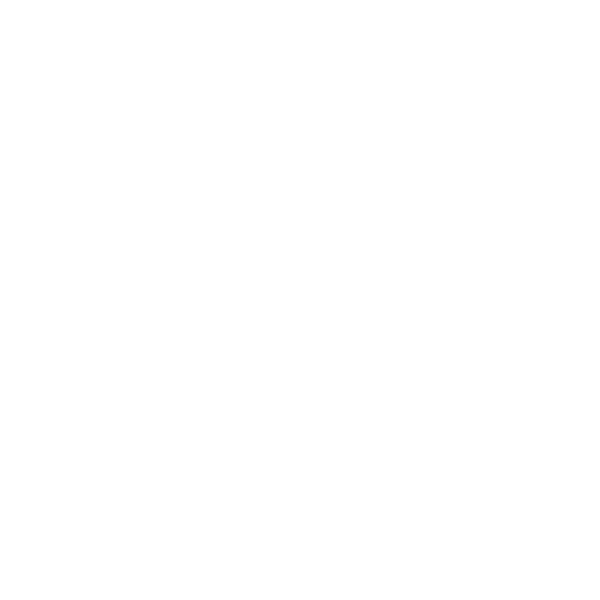}};
   \draw (\xchild+\bx, -\ychild*3+\by) node (kk02) {\includegraphics[width=\fwidth cm]{k.png}};
 
   \draw[arrows={-}, color=black!30, thick] (p01) -- (p11);
   \draw[arrows={-}, color=black!30, thick] (p01) -- (p21);
   \draw[arrows={-}, color=black!30, thick] (p21) -- (k01);
   \draw[arrows={-}, color=black!30, thick] (p21) -- (k11); 
   \draw[arrows={-Latex[length=3*\scale]}, thick] (k11) --(-\xchild/2+\bx , -\ychild*2.5+\by) node[left=1mm] {\tiny $p$} -- (kk01); 
   \draw[arrows={-Latex[length=3*\scale]}, thick] (k11) --(\xchild/2+\bx , -\ychild*2.5+\by) node[right=1mm] {\tiny $p$}-- (kk02);

   \draw (\cx, \cy) node (p02) {\fcolorbox{black}{black}{\includegraphics[width=\fwidth cm]{naked_p0.png}}};
   \draw (\xchild+\cx, -\ychild+\cy) node (p12) {\fcolorbox{red!50}{black}{\includegraphics[width=\fwidth cm]{naked_p1.png}}};
   \draw (-\xchild+\cx, -\ychild+\cy) node (p22){\fcolorbox{black}{black}{\includegraphics[width=\fwidth cm]{naked_p2.png}}};

   \draw  (-\xchild-\xchild+\cx, -\ychild-\ychild+\cy) node (k02) {\fcolorbox{black!30}{black}{\includegraphics[width=\fwidth cm]{naked_p3.png}}};
   \draw  (-\xchild+\xchild+\cx, -\ychild-\ychild+\cy) node (k12) {\fcolorbox{black}{black}{\includegraphics[width=\fwidth cm]{naked_p4.png}}} node[below=4.5mm] {\tiny $v+\mathcal{C}_{\text{path}}$};
 
   \draw[arrows={[length=3*\scale]Latex-}, color=black!30, thick] (p12)-- (p02);
   \draw[arrows={[length=3*\scale]Latex-}, color=black, thick] (p22) -- (p02) ;
   \draw[arrows={[length=3*\scale]Latex-}, color=black!30, thick] (k02) -- (p22) ;
   \draw[arrows={[length=3*\scale]Latex-}, color=black, thick] (k12) -- (p22) ;

   \draw (\dx, \dy) node (p02) {\fcolorbox{black}{black}{\includegraphics[width=\fwidth cm]{naked_p0.png}}};
   \draw (\xchild+\dx, -\ychild+\dy) node (p12) {\fcolorbox{red!50}{black}{\includegraphics[width=\fwidth cm]{naked_p1.png}}};
   \draw (-\xchild+\dx, -\ychild+\dy) node (p22){\fcolorbox{black}{black}{\includegraphics[width=\fwidth cm]{naked_p2.png}}};

   \draw  (-\xchild-\xchild+\dx, -\ychild-\ychild+\dy) node (k02) {\fcolorbox{black!30}{black}{\includegraphics[width=\fwidth cm]{naked_p3.png}}};
   \draw  (-\xchild+\xchild+\dx, -\ychild-\ychild+\dy) node (k12) {\fcolorbox{black}{black}{\includegraphics[width=\fwidth cm]{naked_p4.png}}} node[right=5mm] {\tiny $V$};
 
   \draw[arrows={-Latex[length=3*\scale]}, color=black!30, thick] (p12)-- (p02);
   \draw[arrows={-Latex[length=3*\scale]}, thick] (p22) -- (-\xchild/2+\dx, -\ychild*0.5+\dy) node[left=1mm] {\tiny $Q$} -- (p02) ;
   \draw[arrows={-Latex[length=3*\scale]}, color=black!30, thick] (k02) -- (p22) ;
   \draw[arrows={-Latex[length=3*\scale]}, thick] (k12) -- (-\xchild/2+\dx, -\ychild*1.5+\dy) node[right=1mm] {\tiny $Q$} -- (p22) ;

 \draw (\ax, \ay) node (p0) {\fcolorbox{black}{black}{\includegraphics[width=\fwidth cm]{naked_p0.png}}};
 \draw (\xchild+\ax, -\ychild+\ay) node (p1) {\fcolorbox{red!50}{black!30}{\includegraphics[width=\fwidth cm]{naked_p1.png}}};
 \draw (-\xchild+\ax, -\ychild+\ay) node (p2){\fcolorbox{black}{black!30}{\includegraphics[width=\fwidth cm]{naked_p2.png}}};

 \draw (-\xchild - \xchild+\ax, -\ychild-\ychild+\ay) node (k0) {\fcolorbox{black!30}{black}{\includegraphics[width=\fwidth cm]{naked_p3.png}}};
 \draw (-\xchild + \xchild+\ax, -\ychild-\ychild+\ay) node (k1) {\includegraphics[width=\fwidth cm]{k.png}};
 
 \draw[arrows={-Latex[length=3*\scale]}, color=black!30, thick] (p0) -- (p1);
 \draw[-, rotate around={60:(\xchild/2+\ax, -\ychild/2+\ay)}, color=red] (\xchild/2-\crosslength+\ax, -\ychild/2+\ay) -- (\xchild/2+\crosslength+\ax, -\ychild/2+\ay);
 \draw[-, rotate around={-30:(\xchild/2+\ax, -\ychild/2+\ay)}, color=red] (\xchild/2-\crosslength+\ax, -\ychild/2+\ay) -- (\xchild/2+\crosslength+\ax, -\ychild/2+\ay);
 \draw[arrows={-Latex[length=3*\scale]}, thick] (p0) -- (-\xchild/2+\ax, -\ychild/2+\ay) node[left=1mm] {\tiny $\mathop{\text{argmax}}_a$}-- (p2);
 \draw[arrows={-Latex[length=3*\scale]}, color=black!30, thick] (p2) -- (k0);
 \draw[arrows={-Latex[length=3*\scale]}, thick] (p2) -- (-\xchild + \xchild/2+\ax, -\ychild/2-\ychild+\ay) node[right=1mm] {\tiny $\mathop{\text{argmax}}_a$} -- (k1);
 

   \draw (\xoffset/2.2-0.1, -\yoffset*0.84) node (m) {\fcolorbox{blue!50}{black}{\includegraphics[width=\fwidth cm]{naked_p4.png}}};
   \node[trapezium, fill=black!25, rotate=-90, minimum width=20, outer sep=2pt] (t) at (\xoffset/2.2-0.9, -\yoffset*0.84) {};
   \draw node (ftheta) at (\xoffset/2.2-0.9, -\yoffset*0.84) {\tiny $f_\theta$};
   \draw node (ftheta) at (\xoffset/2.2+0.9, -\yoffset*0.84) {\tiny $=(\boldsymbol{p}, v)$};
 
   \draw[arrows={|-Parenthesis}, yellow, thick] (\xoffset/2.2+0.9, -\yoffset*0.84+0.15) -- (\xoffset/2.2+0.9, -\ychild*2.5) -- (\xchild/2+0.5, -\ychild*2.5);
   \draw[arrows={|-Parenthesis}, blue, thick] (\xoffset/2.2+0.9+0.25, -\yoffset*0.84+0.15) -- (\xoffset/2.2+0.9+0.25, -\ychild*2.5) -- (\xchild/2+\xoffset-0.9, -\ychild*2.5);

 \draw[dashed] (-\xoffset/3.5,\fwidth/2*1.2) -- (\xoffset*0.45, \fwidth/2*1.2) -- (\xoffset*0.45, -\yoffset*0.6) -- (\xoffset/4.5, -\yoffset*0.6) -- (\xoffset/4.5, -\yoffset*0.84) -- (-\xoffset/3.5, -\yoffset*0.84) -- cycle;
 \draw[dashed]  (\xoffset*0.45, -\yoffset*0.6) -- (\xoffset*0.7, -\yoffset*0.6) --  (\xoffset*0.7, -\yoffset*0.84) -- (\xoffset*1.2, -\yoffset*0.84) -- (\xoffset*1.2,\fwidth/2*1.2) -- (\xoffset*0.45, \fwidth/2*1.2);
 \draw[dashed] (-\xoffset/3.5, -\yoffset*0.84) -- (-\xoffset/3.5,-\yoffset*1.75) -- (\xoffset*0.45, -\yoffset*1.75) -- (\xoffset*0.45, -\yoffset*1.05) -- (\xoffset/4.5, -\yoffset*1.05) -- (\xoffset/4.5, -\yoffset*0.84) ;
 \draw[dashed] (\xoffset*0.45, -\yoffset*1.75) -- (\xoffset*1.2, -\yoffset*1.75) -- (\xoffset*1.2, -\yoffset*0.84);
 \draw[dashed] (\xoffset*0.45, -\yoffset*1.05) -- (\xoffset*0.7, -\yoffset*1.05) -- (\xoffset*0.7, -\yoffset*0.84); 

 \draw (-\xoffset/3.5-0.02, -\yoffset*1.75+0.11) node (c0) {} node[right=0.01mm] {\tiny (a) selection};
 \draw (-\xoffset/3.5-0.02, -\yoffset*0.84+0.11) node (c1) {} node[right=0.01mm] {\tiny (b) expansion};
 \draw (\xoffset*0.7-0.02, -\yoffset*0.84+0.11) node (c2) {} node[right=0.01mm] {\tiny (c) simulation};
 \draw (\xoffset*0.45-0.02, -\yoffset*1.75+0.11) node (c3) {} node[right=0.01mm] {\tiny (d) backpropagation};
 \draw (\xoffset/4.5-0.02, -\yoffset*1.05+0.11) node (c4) {} node[right=0.01mm] {\tiny (e) neural network evaluation};

\end{tikzpicture}

\caption{Cycled steps of MCTS in path planning. (a) Iteratively selection tree traversal. (b) Node expansion with policy generator. (c) Simulation with value approximator and cost function $\mathcal{C}_{\text{path}}$. (d) All-way propagation back to root node. (e) Neural network evaluator. }
\label{figurelabel3}
\end{figure}
\end{center}

With the words Monte Carlo in it, MCTS originally works with the idea of randomization within the section of \textit{simulation}. From a \textit{expanded} node, child nodes are visited stochastically until they reach a final 
state, and have an evaluation of such node by then. This method certainly works and will get to a converge when a large number of 
iteration is performed. But it takes a huge amount of time and brings uncertainty to the work, which both are not favored in a real-time system. We would like to use reinforcement learning to get an evaluation neural network 
as a fast rollout policy to boost up the searching process and reduce the randomness at the same time. Before doing that, we need to construct the Monte Carlo tree 
under the automated parking framework.

\subsection{Monte Carlo Tree Search Design}
A tree node $n$ is the combination of a state $s$ and its observation $o$. According to our assumption, with a constant observation $o$, a tree node $n$ has a corresponding state $s$. A tree node $n$ can exist in one of three possible status: \textit{UNEXPLORED}, \textit{EXPLORED}, and \textit{TRIMMED}. A node $n$ is \textit{UNEXPLORED} when it is just spawned from the \textit{expansion} of its parent, and it transitions to the \textit{EXPLORED} status once it is \textit{selected} and \textit{expanded}. A node $n$ is \textit{TRIMMED} if it is trimmed due to certain constraints like collision 
with obstacles, as the red node shown in Fig. 3, or all children of it are trimmed. A tree node $n$ is considered reaching the destination if there is a simple Dubins path\cite{johnson1974application} can be generated connecting its corresponding pose and the corresponding pose of the destination node.

\subsubsection{Selection}
A \textit{selection} will be performed iteratively until it encounters a \textit{UNEXPLORED} node, as is shown in Fig. 3 (a). In order to fully leverage both the prior evaluation and probabilities, the selected action is guided under a designed PUCT algorithm\cite{rosin2011multi}, 
\begin{equation} \label{eqno4}
   \mathop{\text{argmax}}_{a^\prime \in \mathcal{A} }  \ Q(n,a^{\prime}) + C_pP(n,a^\prime)\sqrt{\frac{N(n)+1}{N(n,a^{\prime})+1}}
\end{equation}
where $Q(n, a)$ is the expected return when action $a$ is taken under node $n$, $P(n, a)$ is the policy distribution when action $a$ is taken under node $n$,  $N$ is the visit number of a node or the action under a node, and $C_p$ is an adjustable constant factor to adjust the exploitation preference.

\subsubsection{Expansion}
After \textit{selecting} a node $n$, an immediate \textit{expansion} follows. A uniform sampling strategy is performed within the action space $\mathcal{A}$, generating  
a list of front wheel angle $\delta$ and a pair of bidirectional travel distance $d$ with the same scale, representing forward and reverse action.
Child nodes are therefore spawned from $s$ with the transition function $\mathcal{T}$ under the bicycle kinematic model. A prior probability vector $\boldsymbol{p}$ of each child will be outputted by the neural network estimator $f_\theta$ at the same time, as shown in Fig. 3 (b) and (e). 
If a child node is trimmed by violating certain constraints, its probability share will be split evenly to its living siblings.

\subsubsection{Simulation}
Once $expansion$ is over, we would like to find the value of the selected node $n$ through \textit{simulation}. As what is stated in the \textit{Problem Formulation}, the value(reward) $V$ of a node $n$ can be separated into two parts: the cost it takes from root node to $n$, and a value scalar $v$ denoting the possibility that 
$n$ lies in the best path from the start node to the destination node, which is estimated by the same neural network $f_\theta$ with $\boldsymbol{p}$ at the same time, as what is shown in Fig. 3 (c) and (e),
\begin{equation} \label{eqno5}
   V = \alpha_0 v + \alpha_1 \mathcal{C}_\text{path}
\end{equation}
where $\alpha_0$ and $\alpha_1$ are constant factors which are used to normalize $V$ to $[-1, 1]$, and $\mathcal{C}_\text{path}$ is the cost function.

\subsubsection{Backpropagation}
When the value of a node is obtained, we would like to update the interested properties of its ancestors recursively, as what is shown in Fig. 3 (d), known as \textit{backpropagation}.

\begin{algorithm}[h]
\caption{Backpropagation}
\DontPrintSemicolon
\SetKwFunction{FBack}{Backpropagation}
\SetKwProg{Fn}{Function}{:}{}

\Fn{\FBack{$n, V$}}{
   $V(n) \leftarrow V$ \;
   \While{$n \neq n_0$}{
      \If{$n$ \rm{is connected to destination}} {
         $V \leftarrow \textbf{Max}(V, V(n))$
      }
      $n, a \leftarrow \text{parent of} \ n , \text{corresponding action}$ \;
      $Q(n, a) \leftarrow \frac{N(n, a)Q(n, a) + V}{N(n,a)+1}$ \;
      $N(n, a) \leftarrow N(n,a)+1$ \;
      $N(n) \leftarrow N(n) + 1$
      
   }
}

\end{algorithm}
As shown in Alg. 1, $V(n)$ is the stored estimated value of a node $n$, and $n_0$ is the root node. We would like to keep the best estimation 
if a node is connected to the destination as the value propagated back to enhance the power of successful prior knowledge.

The MCTS algorithm is terminated if it meets one of the following conditions: (1) the tree has fully expanded; (2) it reaches a time limit or a visited node limit; 
(3) a certain number of paths is found.

\subsection[short]{Reinforcement Learning Design}
MCTS itself can be treated as a natural \textit{policy improvement} operator. Given prior value estimations, it gradually updates its selection policy.
After each MCTS is terminated, retrieving the information from the search tree will help us re-estimate the value, as \textit{policy evaluation}. By iteratively 
executing \textit{policy improvement} and \textit{policy evaluation} process, we have a \textit{policy iteration} framework where MCTS takes part in both processes.

\begin{center}
   \begin{figure}[thpb]
      \centering
\begin{tikzpicture}[]
   \tikzmath{\scale = 1.1; \xoffset = 0.05*\scale;\xchild = 0.7*\scale;\ychild=1.6*\scale;\fwidth=1.0*\scale;\yoffset=0.05*\scale;}
   \tikzmath{\crosslength = 0.1*\scale;}
   \tikzmath{\xbig = 2*\scale; \ybig=0.6*\scale; \yc=0.12*\scale; \yt=1.5*\scale;}
   \tikzmath{\ax=0.0; \ay=-\yoffset; \bx=0.0; \by=0.0; \cx=\xoffset; \cy=0.0; \dx=\xoffset; \dy=-\yoffset;}
   \tikzmath{\ymcts=6;} 
  \node[scale=0.6] at (2.5, -2.1+\ymcts) {\scriptsize \textbf{MCTS}};
  \node (rect) at (2.5, -3.3+\ymcts) [draw, fill=yellow!20, thick, rounded corners=5pt, minimum width=2.0 cm, minimum height = 2.0cm, outer sep=4pt] {};

  \node[draw, circle, minimum size=0.2cm, inner sep=0pt, fill=black!30, thick] (a) at (2.5, -2.5+\ymcts) {};
  \node[draw=black!30, circle, minimum size=0.2cm, inner sep=0pt, fill=white] (b) at (1.75, -3+\ymcts) {};
  \node[text=black!30] (c) at (2.25, -3+\ymcts) {...};
  \node[draw, circle, minimum size=0.2cm, inner sep=0pt,  fill=black!30, thick] (d) at (2.75, -3+\ymcts) {};
  \node[draw=black!30, circle, minimum size=0.2cm, inner sep=0pt, fill=white] (e) at (3.25, -3+\ymcts) {};
  \node[text=black!30] (f) at (1.75, -3.5+\ymcts) {...};
  \node[draw, circle, minimum size=0.2cm, inner sep=0pt,  fill=black!30, thick] (g) at (2.5, -3.5+\ymcts) {};
  \node[draw=black!30, circle, minimum size=0.2cm, inner sep=0pt, fill=white] (h) at (3, -3.5+\ymcts) {};
  \node[draw, circle, minimum size=0.2cm, inner sep=0pt,  fill=black!30, thick] (i) at (2.5, -4+\ymcts) {};

  \begin{scope}[>=Stealth]
  \draw[->, color=black!30] (a)--(b);
  \draw[->, color=black!30] (b)--(f);
  \draw[->, thick] (d)--(a);
  \draw[->, color=black!30] (a)--(e);
  \draw[->, color=black!30] (d)--(h);
  \draw[->, thick] (i)--(g);
  \draw[->, thick] (g)--(d);
  \end{scope}
   \node[opacity=0.2] at (0, 0) {\fcolorbox{black}{black}{\includegraphics[width=\fwidth cm]{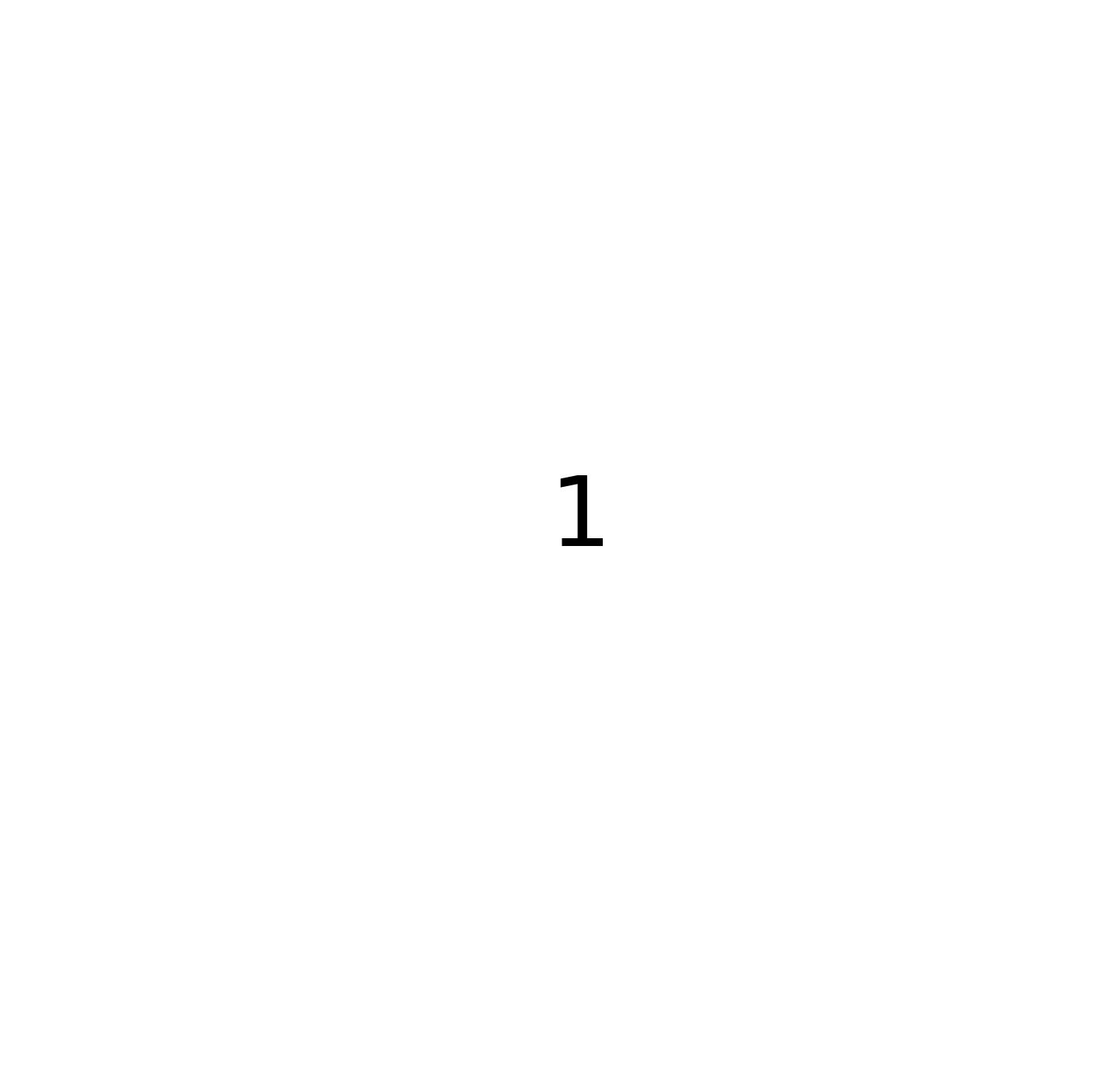}}};
   \node[opacity=0.2] at (\xoffset, -\yoffset) {\fcolorbox{black}{black}{\includegraphics[width=\fwidth cm]{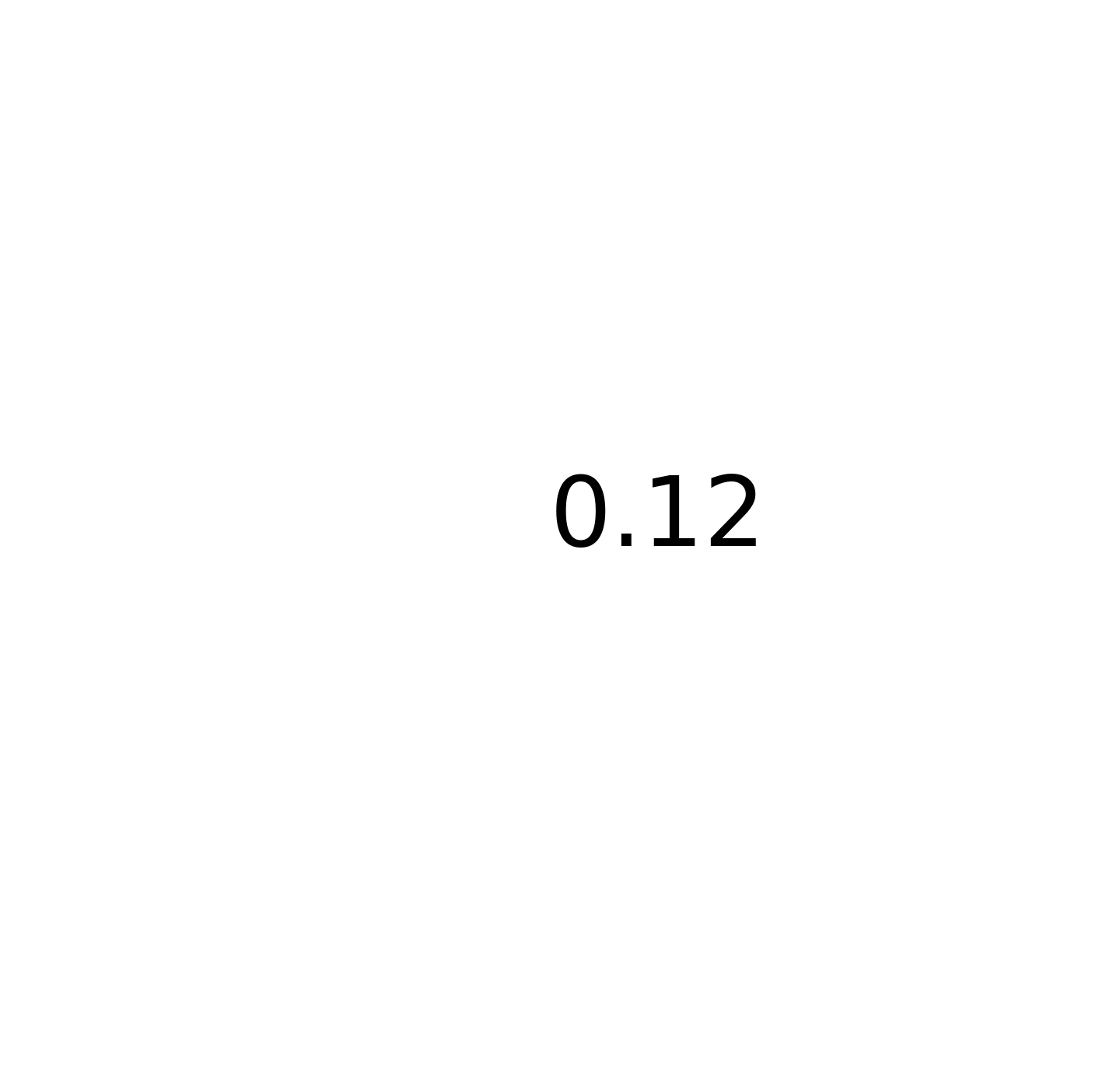}}};
   \node[opacity=0.2] at (2*\xoffset, -2*\yoffset) {\fcolorbox{black}{black}{\includegraphics[width=\fwidth cm]{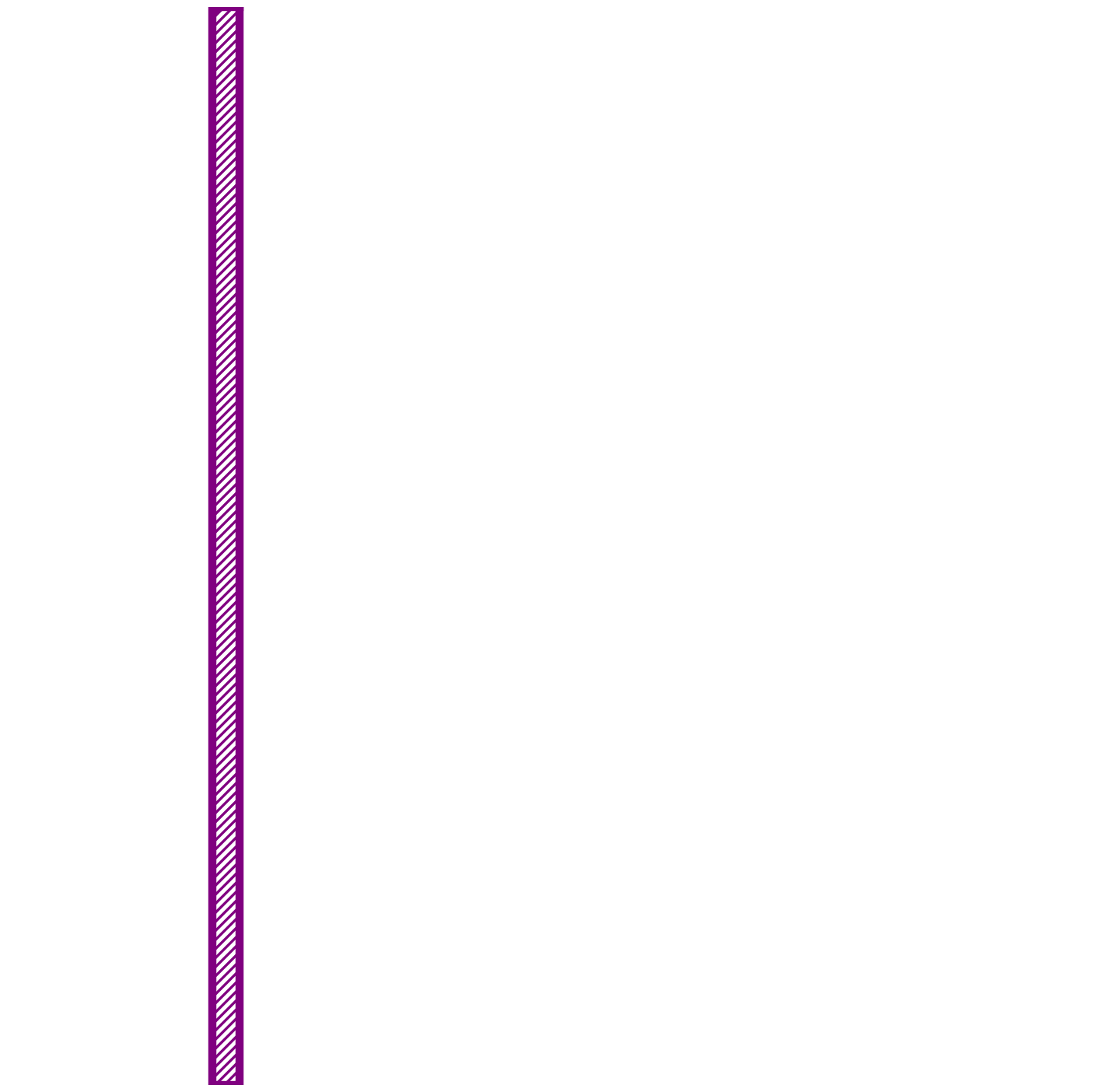}}};
   \node[opacity=0.2] at (3*\xoffset, -3*\yoffset) {\fcolorbox{black}{black}{\includegraphics[width=\fwidth cm]{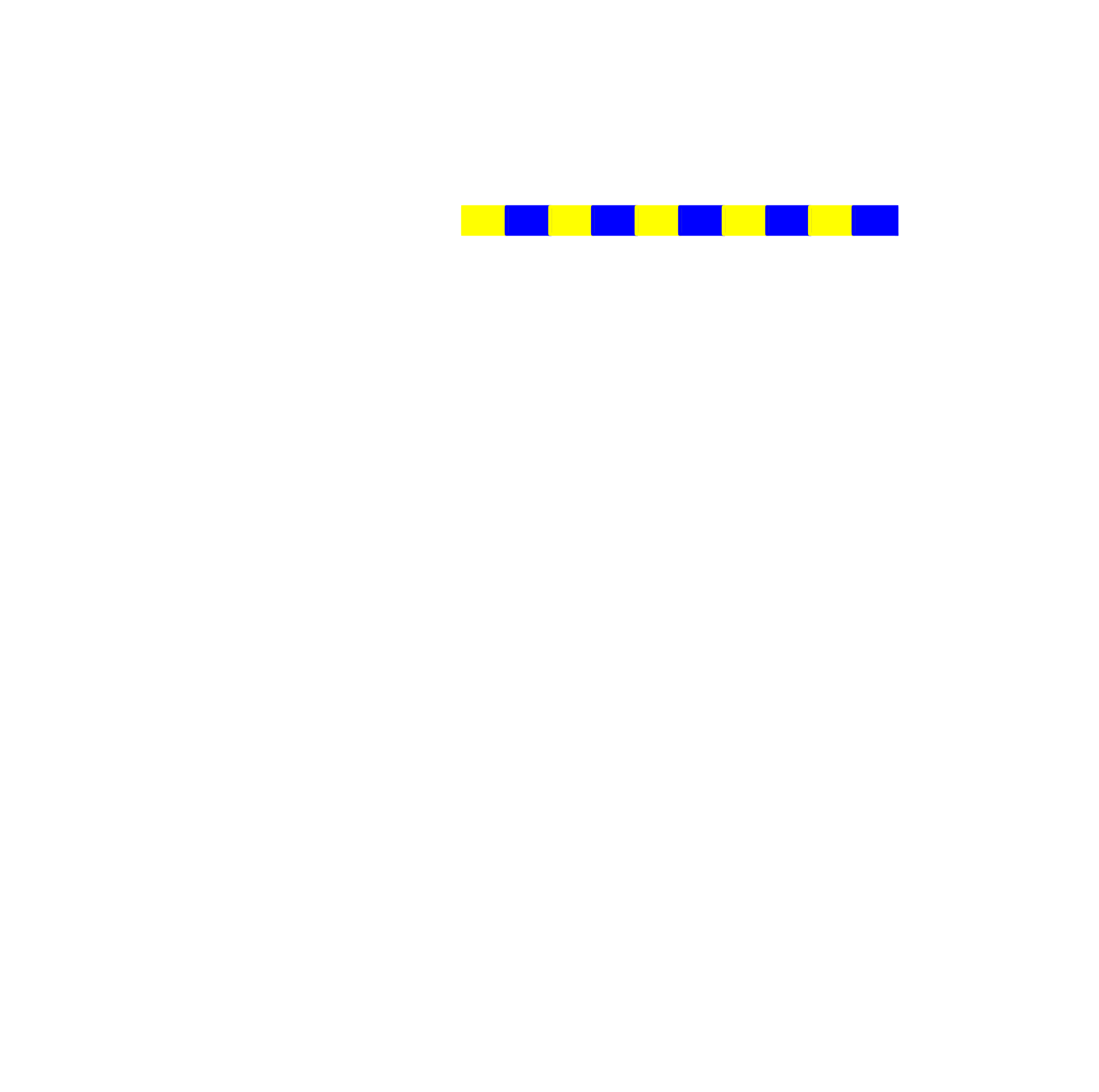}}};
   \node[opacity=0.0] (input) at (6*\xoffset, -3.5*\yoffset) {\fcolorbox{black}{black}{\includegraphics[width=\fwidth cm]{input_bump.png}}};
   \node[opacity=0.2] at (4*\xoffset, -4*\yoffset) {\fcolorbox{black}{black}{\includegraphics[width=\fwidth cm]{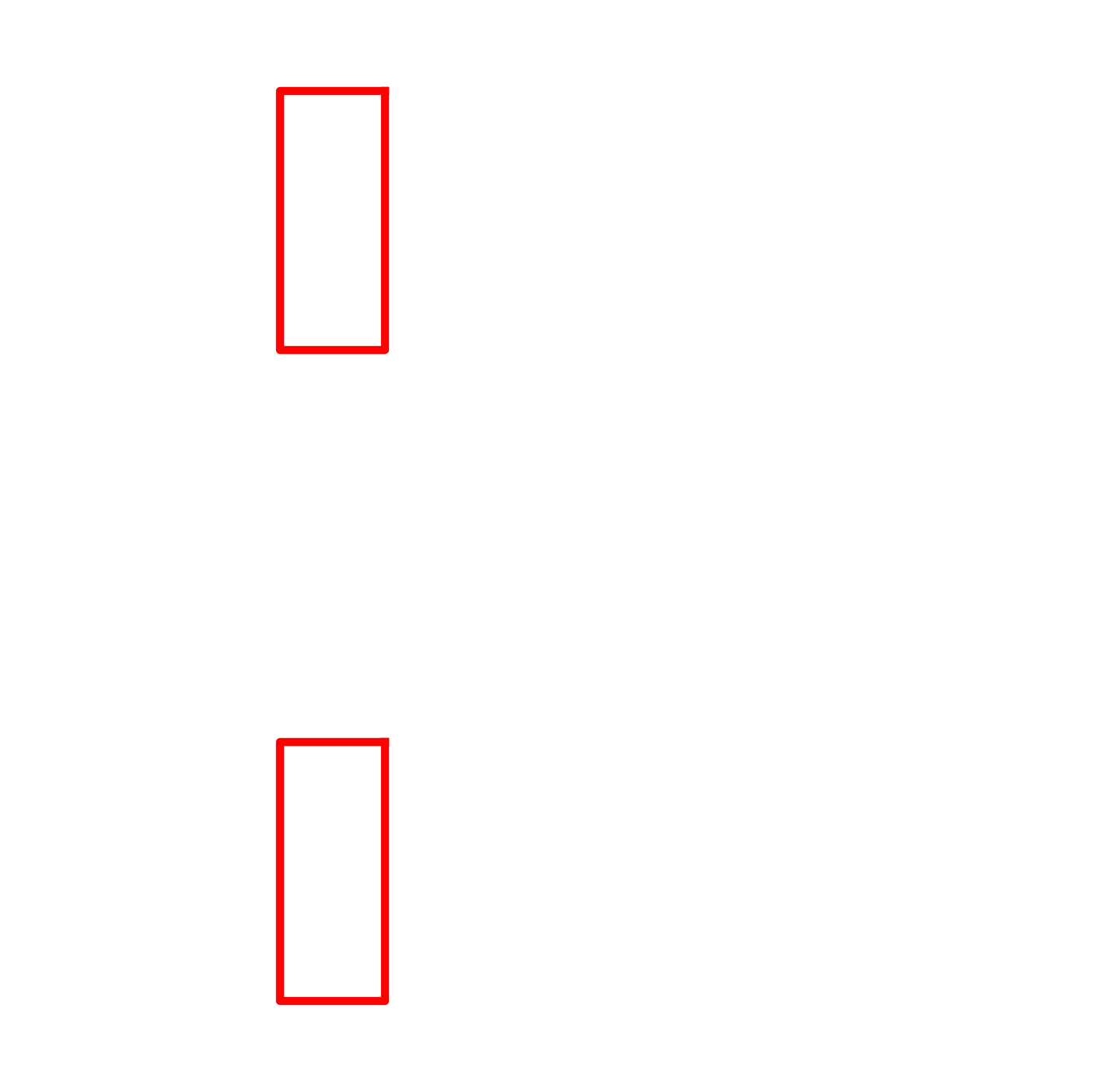}}};
   \node[opacity=0.2] at (7*\xoffset, -7*\yoffset) {\fcolorbox{black}{black}{\includegraphics[width=\fwidth cm]{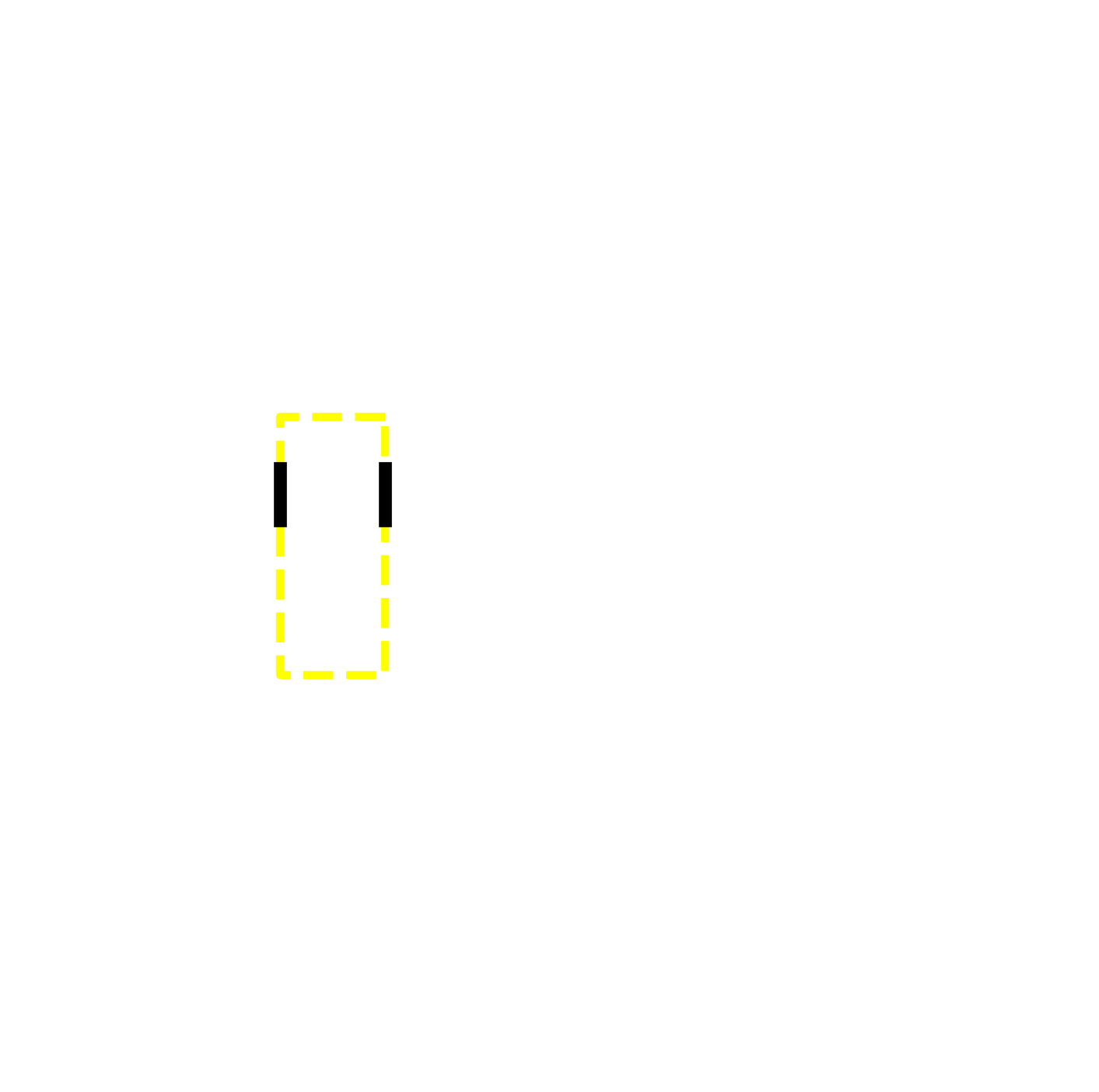}}};
   \node[opacity=0.2] at (5*\xoffset, -5*\yoffset) {\fcolorbox{black}{black}{\includegraphics[width=\fwidth cm]{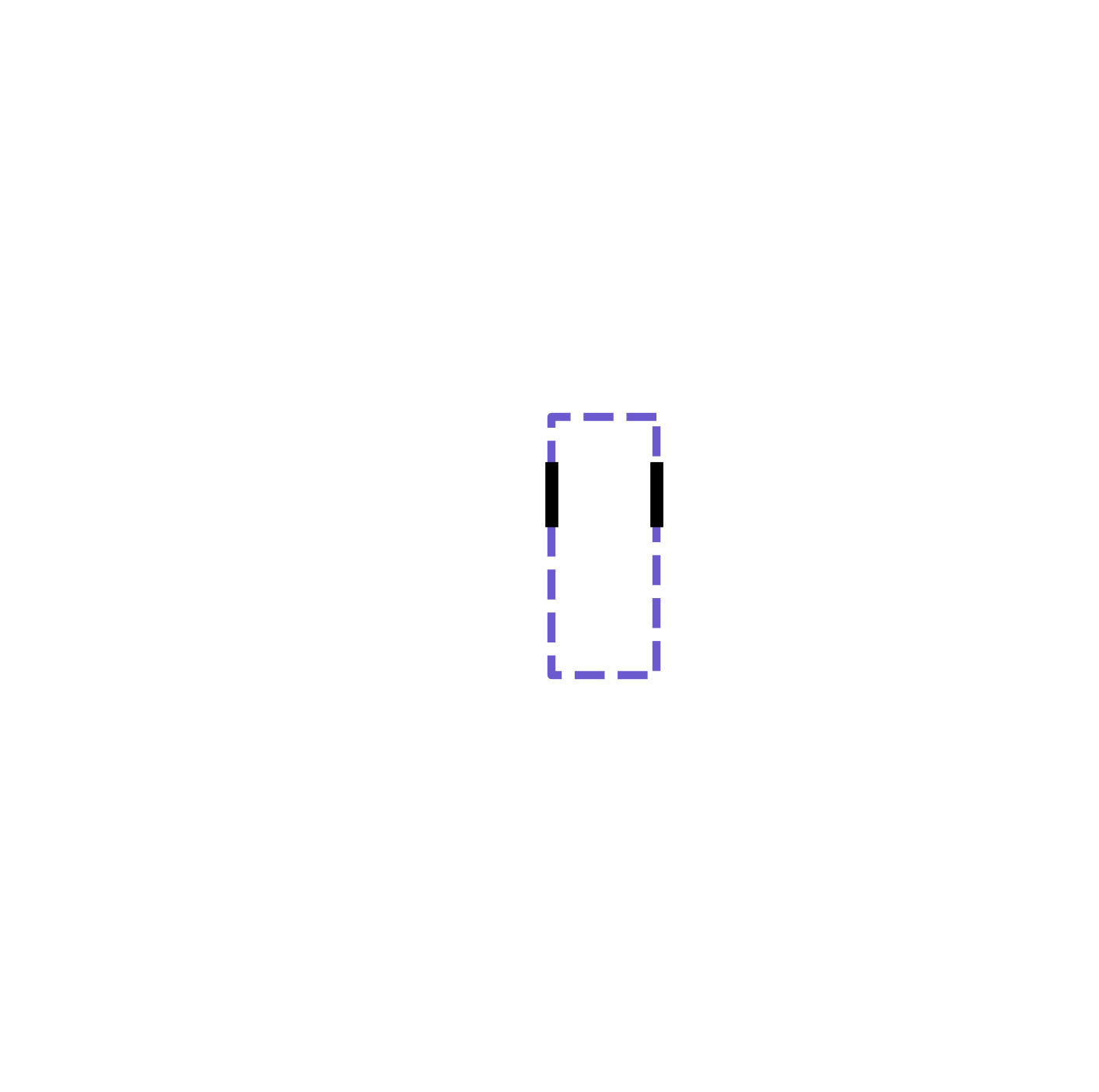}}};
   \node[opacity=0.2] at (6*\xoffset, -6*\yoffset) {\fcolorbox{black}{black}{\includegraphics[width=\fwidth cm]{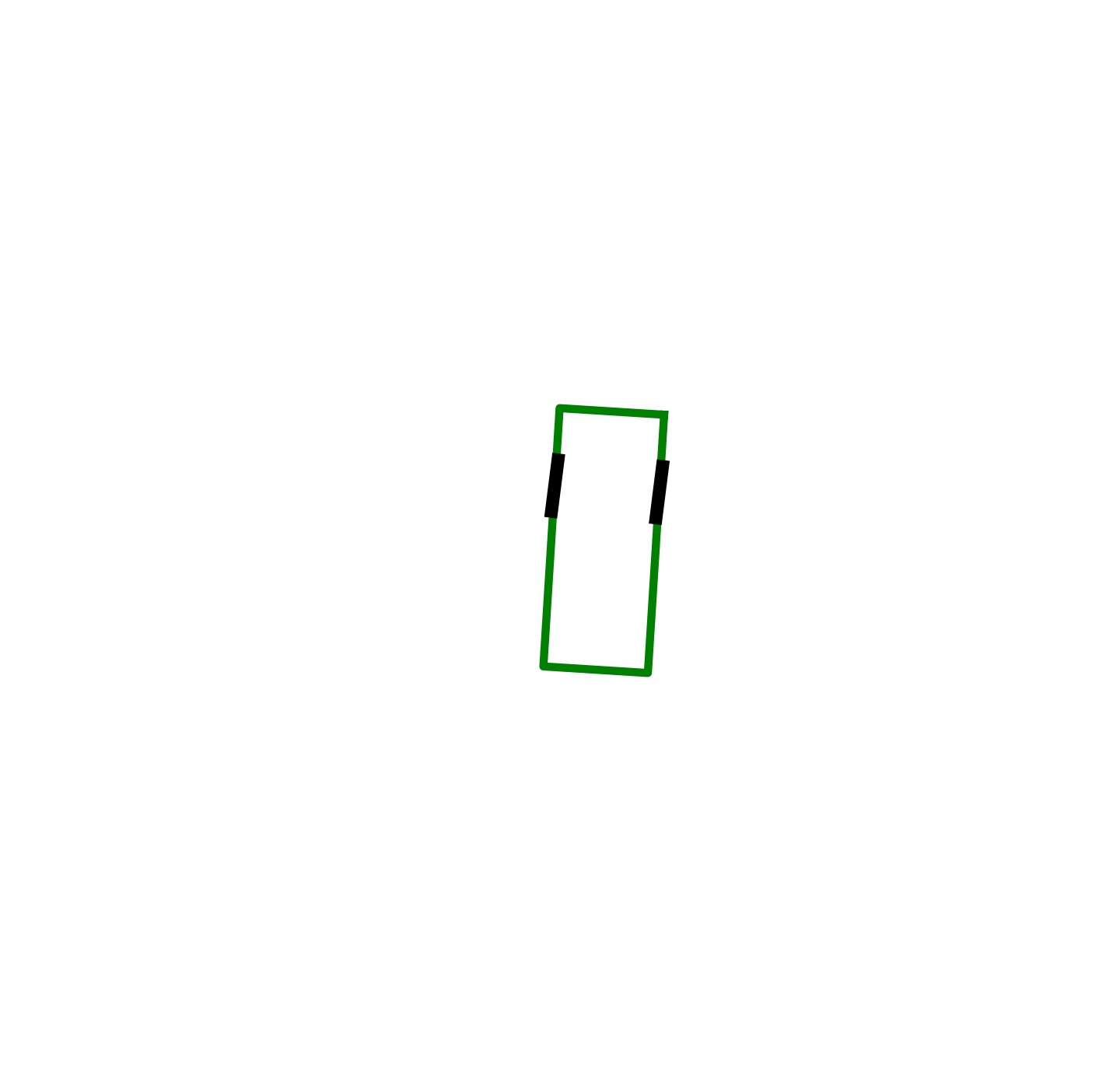}}};
 
   \node[trapezium, fill=black!25!blue!10, rotate=-90, minimum width=30, outer sep=2pt] (bb) at (\xbig, -3.5*\yoffset) {};
 
   \node[rectangle, fill=black!25, minimum width=3, minimum height=50, inner sep=0, outer sep=2pt] (f) at (\xbig*1.6, -3.5*\yoffset) {};
 
   \node[opacity=0.0, circle, fill=black!25!blue!10, minimum width=2, inner sep=0, outer sep=2pt] (h1l) at (\xbig*2.15, -3.5*\yoffset+\ybig) {};
   \node[opacity=0.0, circle, fill=black!25!blue!10, minimum width=2, inner sep=0, outer sep=2pt] (h1r) at (\xbig*2.15+\yc, -3.5*\yoffset+\ybig) {};
   \node[circle, fill=black!25!blue!10, minimum width=2, inner sep=0] (h1-1) at (\xbig*2.15, -3.5*\yoffset+\ybig-\yc) {};
   \node[circle, fill=black!25!blue!10, minimum width=2, inner sep=0] (h1-2) at (\xbig*2.15, -3.5*\yoffset+\ybig+\yc) {};
   \node[circle, fill=black!25!blue!10, minimum width=2, inner sep=0] (h1-3) at (\xbig*2.15+\yc, -3.5*\yoffset+\ybig) {};
   \node[circle, fill=black!25!blue!10, minimum width=2, inner sep=0] (h1-4) at (\xbig*2.15+\yc, -3.5*\yoffset+\ybig+2*\yc) {};
   \node[circle, fill=black!25!blue!10, minimum width=2, inner sep=0] (h1-5) at (\xbig*2.15+\yc, -3.5*\yoffset+\ybig-2*\yc) {};
   \draw[-,color=black!80] (h1-1) -- (h1-3);
   \draw[-,color=black!80] (h1-1) -- (h1-4);
   \draw[-,color=black!80] (h1-1) -- (h1-5);
   \draw[-,color=black!80] (h1-2) -- (h1-3);
   \draw[-,color=black!80] (h1-2) -- (h1-4);
   \draw[-,color=black!80] (h1-2) -- (h1-5);

   \node[opacity=0.0, circle, fill=black!25!blue!10, minimum width=2, inner sep=0,outer sep=2pt] (h2l) at (\xbig*2.15, -3.5*\yoffset-\ybig) {};
   \node[opacity=0.0, circle, fill=black!25!blue!10, minimum width=2, inner sep=0,outer sep=2pt] (h2r) at (\xbig*2.15+\yc, -3.5*\yoffset-\ybig) {};
   \node[circle, fill=black!25!blue!10, minimum width=2, inner sep=0] (h2-1) at (\xbig*2.15, -3.5*\yoffset-\ybig-\yc) {};
   \node[circle, fill=black!25!blue!10, minimum width=2, inner sep=0] (h2-2) at (\xbig*2.15, -3.5*\yoffset-\ybig+\yc) {};
   \node[circle, fill=black!25!blue!10, minimum width=2, inner sep=0] (h2-3) at (\xbig*2.15+\yc, -3.5*\yoffset-\ybig) {};
   \node[circle, fill=black!25!blue!10, minimum width=2, inner sep=0] (h2-4) at (\xbig*2.15+\yc, -3.5*\yoffset-\ybig+2*\yc) {};
   \node[circle, fill=black!25!blue!10, minimum width=2, inner sep=0] (h2-5) at (\xbig*2.15+\yc, -3.5*\yoffset-\ybig-2*\yc) {};
   \draw[-,color=black!80] (h2-1) -- (h2-3);
   \draw[-,color=black!80] (h2-1) -- (h2-4);
   \draw[-,color=black!80] (h2-1) -- (h2-5);
   \draw[-,color=black!80] (h2-2) -- (h2-3);
   \draw[-,color=black!80] (h2-2) -- (h2-4);
   \draw[-,color=black!80] (h2-2) -- (h2-5);

   \node[rectangle, fill=black!25, minimum width=3, minimum height=15, inner sep=0, outer sep=2pt] (p) at (\xbig*2.7, -3.5*\yoffset+\ybig) {};
   \node[opacity=0] (pp) at (\xbig*2.7, -3.5*\yoffset+\ybig+0.22) {};
   \node (tp) at (\xbig*2.7+0.2, -3.5*\yoffset+\ybig) {\tiny $\boldsymbol{p}$};
 
   \node[rectangle, fill=black!25, minimum width=3, minimum height=3, inner sep=0, outer sep=2pt] (v) at (\xbig*2.7, -3.5*\yoffset-\ybig) {};
   \node (tv) at (\xbig*2.7+0.2, -3.5*\yoffset-\ybig) {\tiny $v$};

   \draw[arrows={-Latex[length=3*\scale]}, color=black] (input) -- (bb);
   \draw[arrows={-Latex[length=3*\scale]}, color=black] (bb) -- (f);
   \draw[arrows={-Latex[length=3*\scale]}, color=black] (f) -- (h1l);
   \draw[arrows={-Latex[length=3*\scale]}, color=black] (f) -- (h2l);
   \draw[arrows={-Latex[length=3*\scale]}, color=black] (h1r) -- (p);
   \draw[arrows={-Latex[length=3*\scale]}, color=black] (h2r) -- (v);

   \node[scale=0.8] (t1) at (3.5*\xoffset, -\yt) {\scriptsize Input};
   \node[scale=0.8] (t2) at (\xbig, -\yt) {\scriptsize Backbone};
   \node[scale=0.8] (t3) at (\xbig*1.6, -\yt) {\scriptsize Feature};
   \node[scale=0.8] (t4) at (\xbig*2.15, -\yt) {\scriptsize Head};
   \node[scale=0.8] (t5) at (\xbig*2.7, -\yt) {\scriptsize Target};

  \draw[->, dashed, thick, green!20!red] (rect) -- (input);
  \draw[->, dashed, thick, green!20!red] (pp) -- (rect);

  \draw[dashed] (-1,1.2) -- (6.5, 1.2) -- (6.5, 4.2) -- (-1,4.2) -- cycle;
  \draw[dashed] (6.5, 1.2) -- (6.5, -2.0) -- (-1,-2.0) -- (-1,1.2);
  
  \draw (-1, 4.2-0.15) node (ca) {} node[right=0.01mm] {\tiny (a) policy improvement};
  \draw (-1, 1.2-0.15) node (cb) {} node[right=0.01mm] {\tiny (b) policy evaluation};
\end{tikzpicture}

\caption{Reinforcement learning pipeline. (a) Training data is retrieved once MCTS is terminated, as the \textit{policy improvement} finished. (b) The architecture of the \textit{evaluation} neural network is composed of a convolutional backbone and two separated MLP heads. The updated model parameters are used in the next MCTS iteration.}
\label{figurelabel4}
\end{figure}
\end{center}

A neural network is used to project a given state to an estimated policy and value. It will be trained against the assembled labels from MCTS outcomes. 
The general architecture of this network is shown in Fig. 4. The input of the neural network is a stacked tensor consisting of the following layers: (1) the occupancy layers for observation space like different types of obstacles; (2) the occupancy layers for vehicle agent in the current state, parent state and destination state; 
(3) the numeric layers for other interested current and previous state's properties like gear and steering wheel angle.

The input tensor is fed into a \textit{backbone} consisting of several convolutional blocks with corresponding batch normalization layer and activation layer.
A single dimension hidden \textit{feature} layer is outputted by \textit{backbone}, and goes to two separate multilayer perceptron \textit{heads}, which project the hidden \textit{feature} into 
the \textit{policy} vector and the \textit{value} scalar respectively.

Given a fair amount of time, feasible paths would be found if they exist. After the search process is terminated, 
we would have a tree with nodes that lie in feasible paths as good nodes and the out-lied ones as bad nodes. We use the Farthest Point Sampling\cite{qi2017pointnet++} strategy to sample the same amount of both good and bad nodes for positive and negative data.

For a sampled node $n$, we would like to use 
\begin{equation} \label{eqno6}
   \pi(a|n) = \frac{N(n,a)^{\frac{1}{\tau}}}{\sum_b N(n,b)^{\frac{1}{\tau}}}
\end{equation}
as the policy label, where $\tau$ is the temperature factor which is used to tune the confidence for the next \textit{policy evaluation}. We would like to label the good node as $r=1$ and the bad node as $r=0$, representing the chance if there is a good path from the node to the destination.

The network parameter $\theta$ is then trained against a loss function that simply sums up the cross-entropy loss of policy \textit{head} and the mean-square loss of value \textit{head}, as
\begin{equation} \label{eqno7}
   \mathop{\text{argmin}}_{\theta}  \ -\boldsymbol{\pi}^{\text{T}}\log\boldsymbol{p} + (v-r)^2
\end{equation}

The network is trained after each round of MCTS, and its updated parameters are used for the next round of MCTS exploration, as what is shown in Fig. 1.

\section{EXPERIMENTS}

The training and inference experiments are evaluated on a computer with 32GB memory, an AMD Ryzen 7 3700x CPU, and an Nvidia GeForce RTX 3060 GPU.

Derived from real road test scenarios, we constructed thousands of simulated parking scenarios in different types, enumerating the critical features one could have like the size of the parking spots, obstacles’ poses, agent's starting poses, etc.
The generated scenarios are separated into training, test and validation datasets with a ratio of 70\%:15\%:15\%. 

The performance of the MCTS in the inference stage is evaluated based on the planning time it takes to generate a feasible path that meets the path generated by the Hybrid A* \cite{dolgov2008practical} algorithm in its quality. We recorded the behavior of Hybrid A* and MCTS at different training steps through the validation dataset, as what is shown in Fig. 5 (a). The discretization setup is 0.1 m for the position and 0.01 Rad for the orientation.
The planning time MCTS takes to find a path drops quickly in the opening stages, and gradually converges to a stable condition. In the end, the median planning time required by MCTS through the dataset is only 7.2\% compared to the time taken by Hybrid A*. The performance of trained MCTS and Hybrid A* under multiple discretizations is shown in Fig. 5 (b), which suggests that model can be easily scaled up to different setups.

The algorithm has been deployed on autonomous vehicles for daily operations. We selected three distinct parking scenarios from real road test data for analysis. The path planning is tailored to initiate from the parking spot towards the vehicle's current position to ensure rapid convergence.

\begin{center}
\begin{figure}[thpb]
    \centering
    \begin{tikzpicture}[]
    \tikzmath{\sszie = 4.2;}
    \draw (-\sszie*0.5, 0) node (p01)  {\includegraphics[width=\sszie cm]{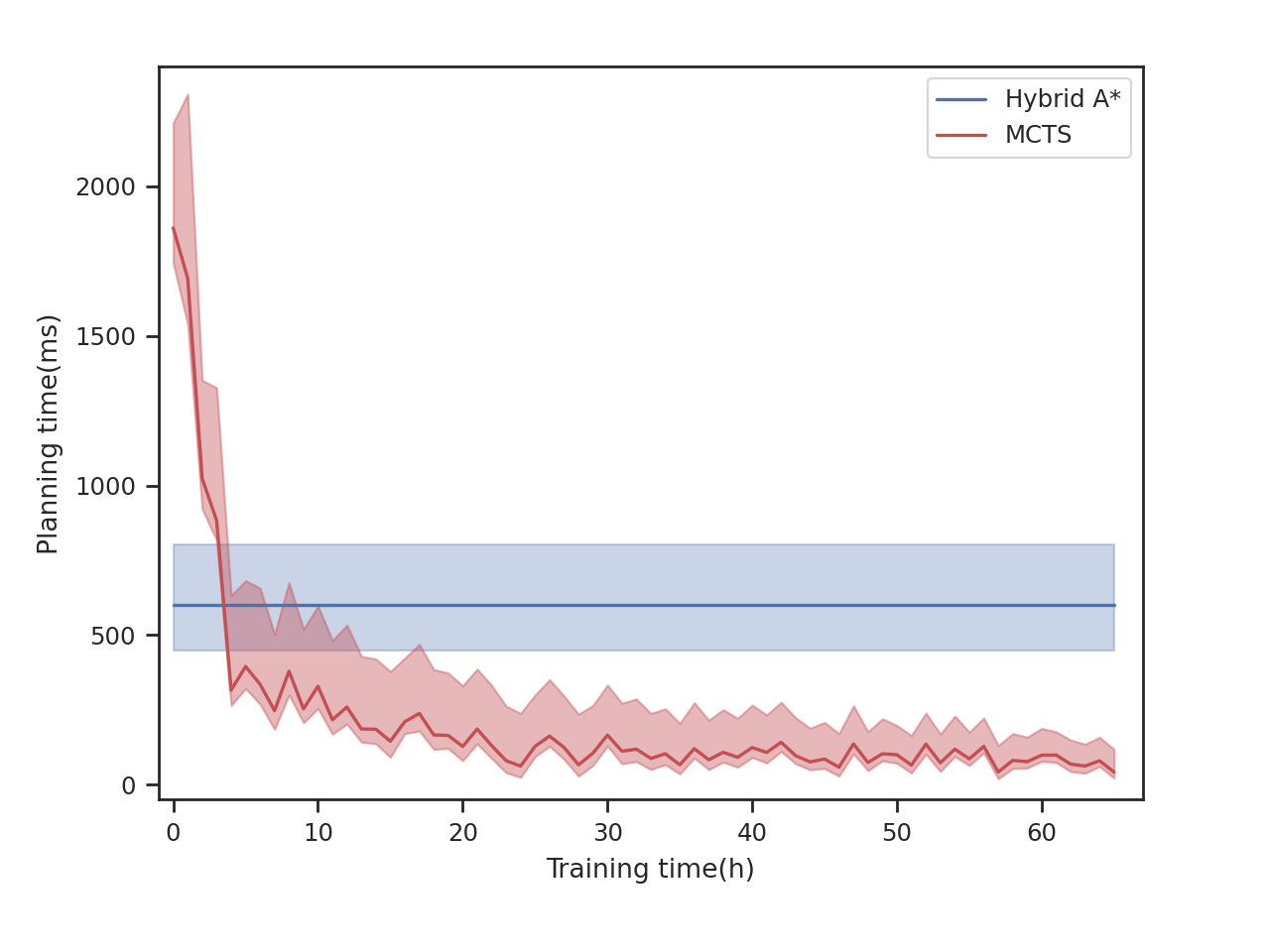}};
    \draw (\sszie*0.5-0.1, 0) node (p02)  {\includegraphics[width=\sszie cm]{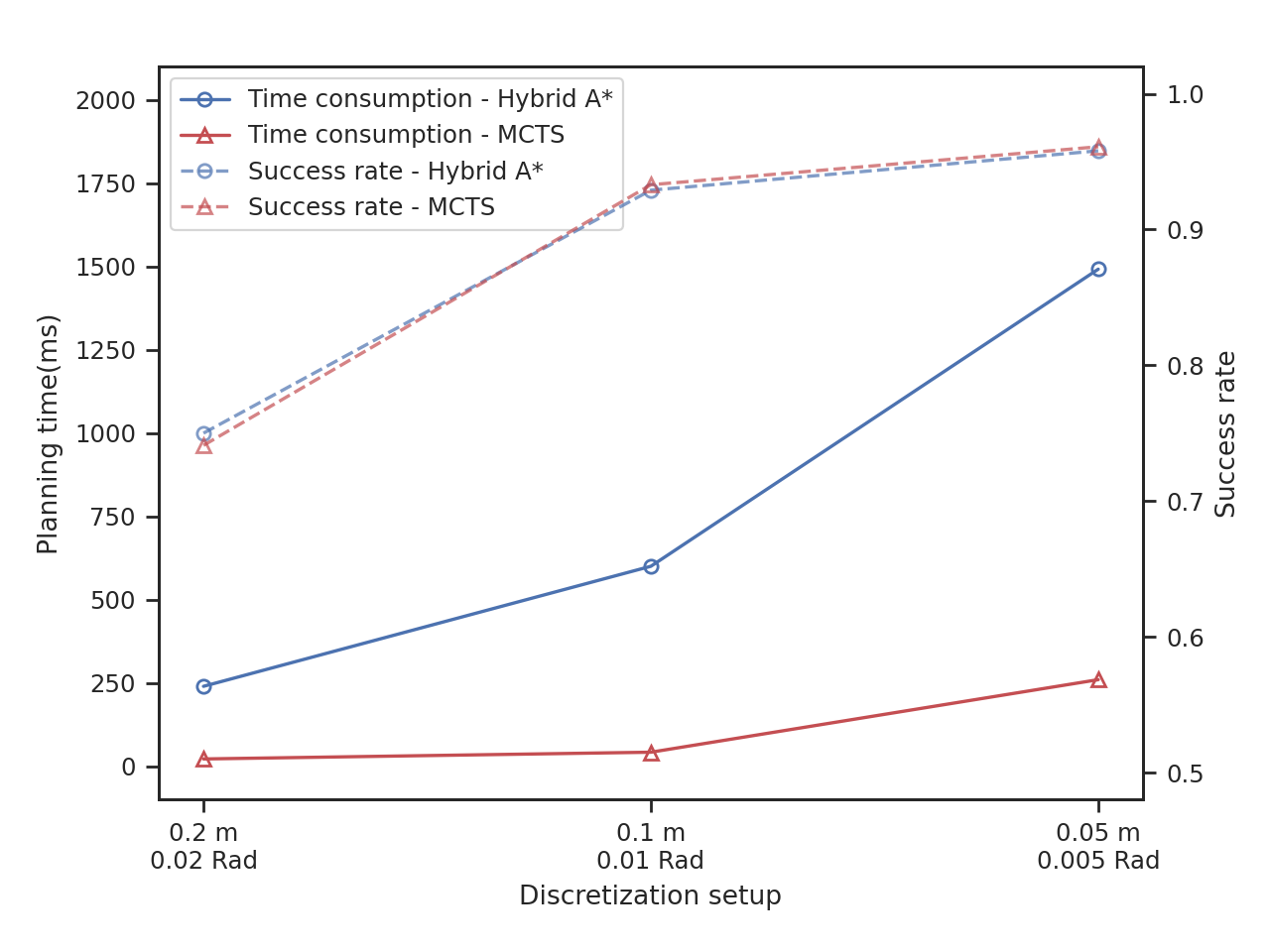}};

    \node[] at (-\sszie*0.5, -1.7) {\scriptsize (a)};
    \node[] at (\sszie*0.5-0.1, -1.7) {\scriptsize (b)};
   \end{tikzpicture}
\caption{(a) Training metric. The darker line is the median over the validation dataset, and the pale shaded area is formed by the 10th and 90th percentiles. The blue one is the planning time which Hybrid A* algorithm takes while the red one is the performance of MCTS at different training steps. (b) Planning time and success rate comparison between trained MCTS and Hybrid A* under different discretization setups over the validation dataset.}
\label{figurelabel5}
\end{figure}
\end{center}

\begin{center}
   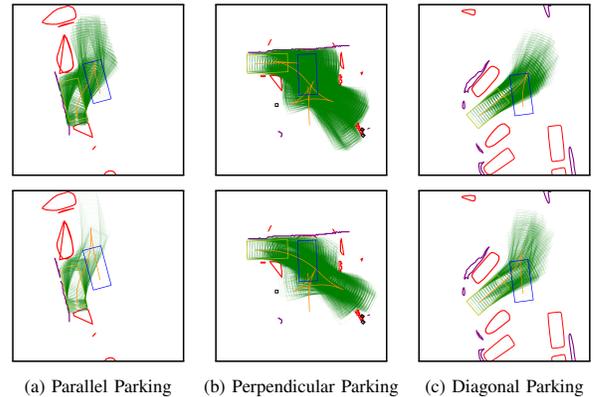
\begin{figure}[thpb]
      \centering
      \begin{tikzpicture}[]
         \tikzmath{\scale = 0.45; \xoffset = 6*\scale;\xchild = 0.7*\scale;\ychild=1.6*\scale;\fwidth=5.0*\scale;\yoffset=5.5*\scale;}
         \tikzmath{\crosslength = 0.1*\scale;}
         \tikzmath{\ax=0.0; \ay=-\yoffset; \bx=0.0; \by=0.0; \cx=\xoffset; \cy=0.0; \dx=\xoffset; \dy=-\yoffset;}
         
         \draw (0, 0) node (p01) {\fcolorbox{black}{black}{\includegraphics[width=\fwidth cm]{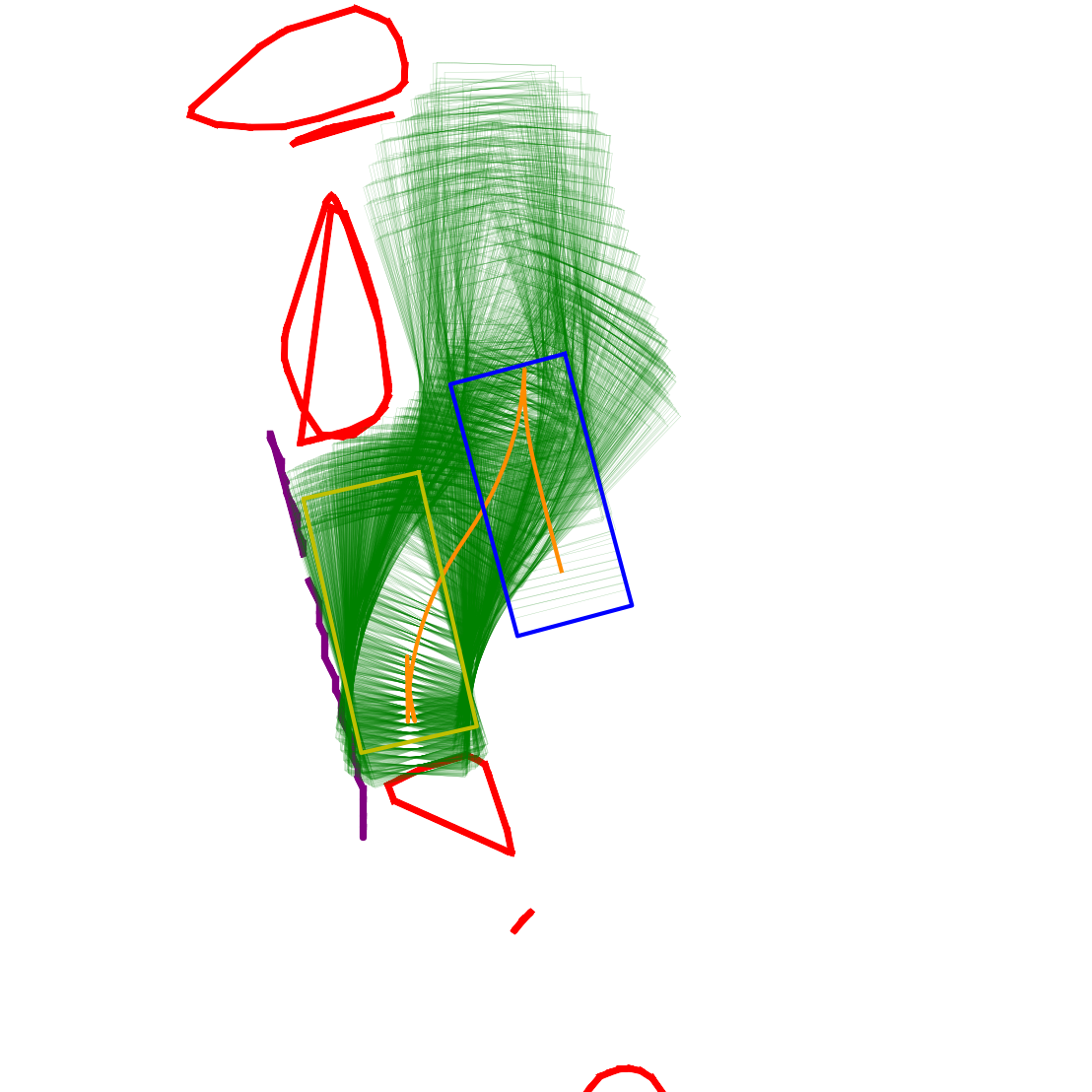}}};
         \draw (0, -\yoffset) node (p02) {\fcolorbox{black}{black}{\includegraphics[width=\fwidth cm]{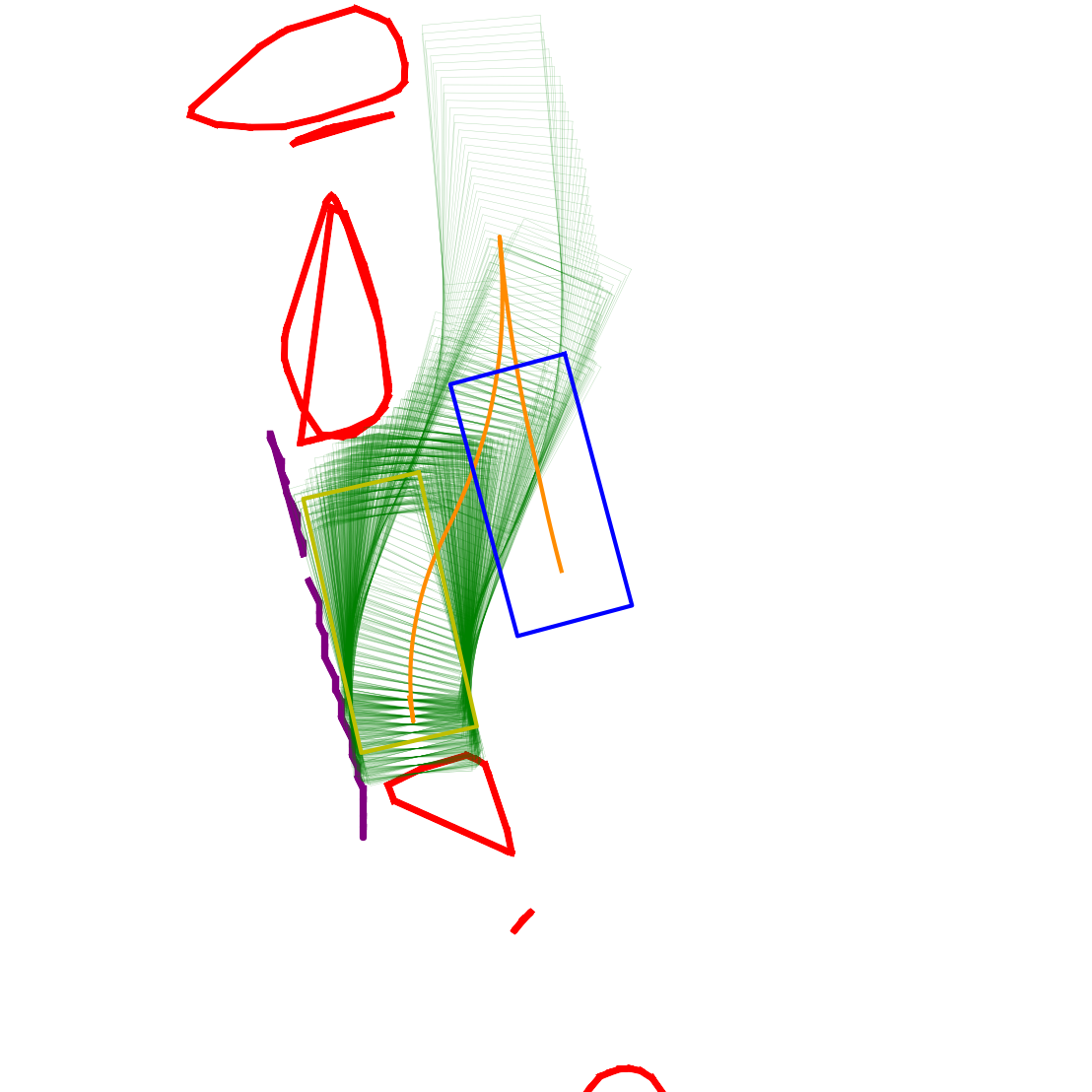}}};

         \draw (\xoffset, 0) node (p11) {\fcolorbox{black}{black}{\includegraphics[width=\fwidth cm]{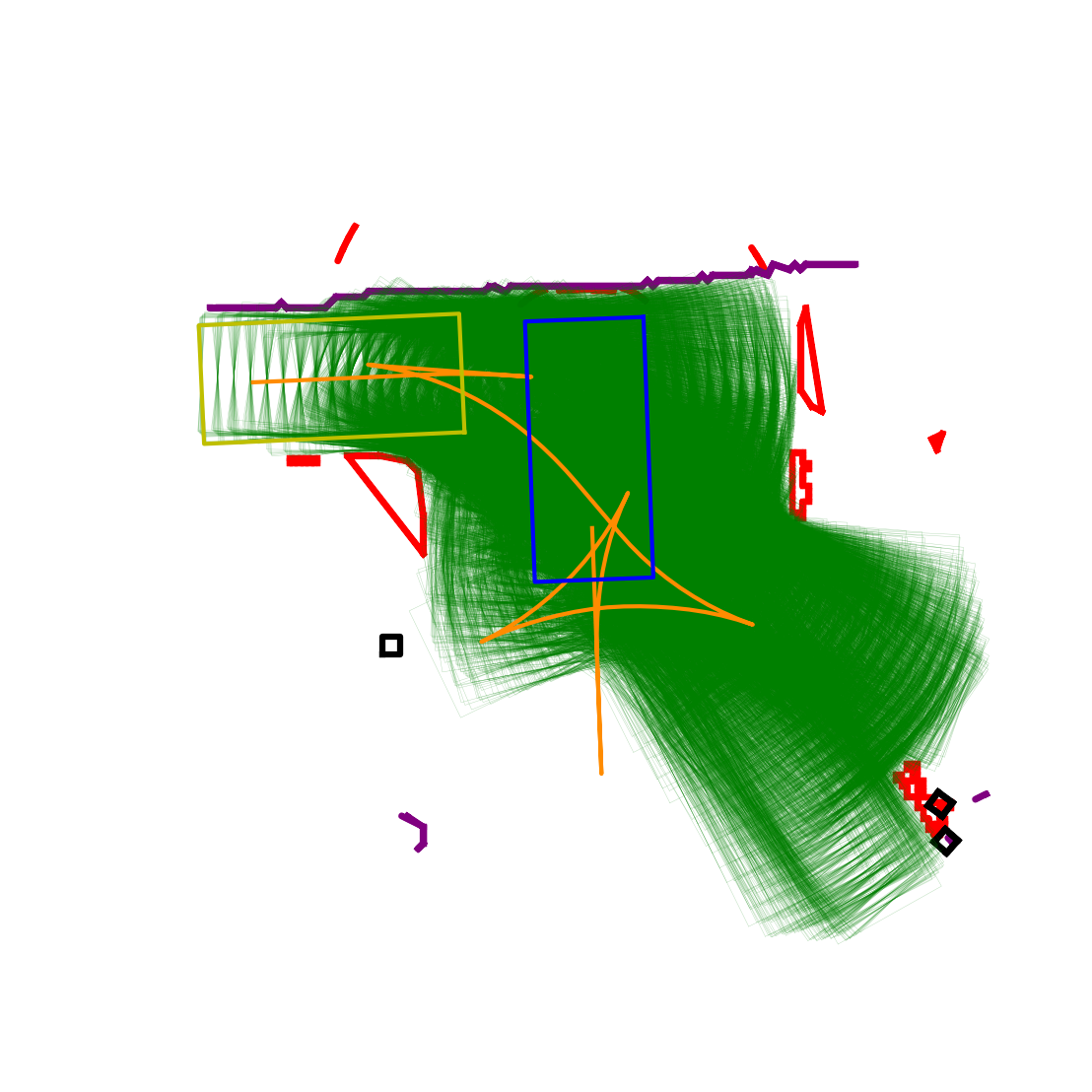}}};
         \draw (\xoffset, -\yoffset) node (p12) {\fcolorbox{black}{black}{\includegraphics[width=\fwidth cm]{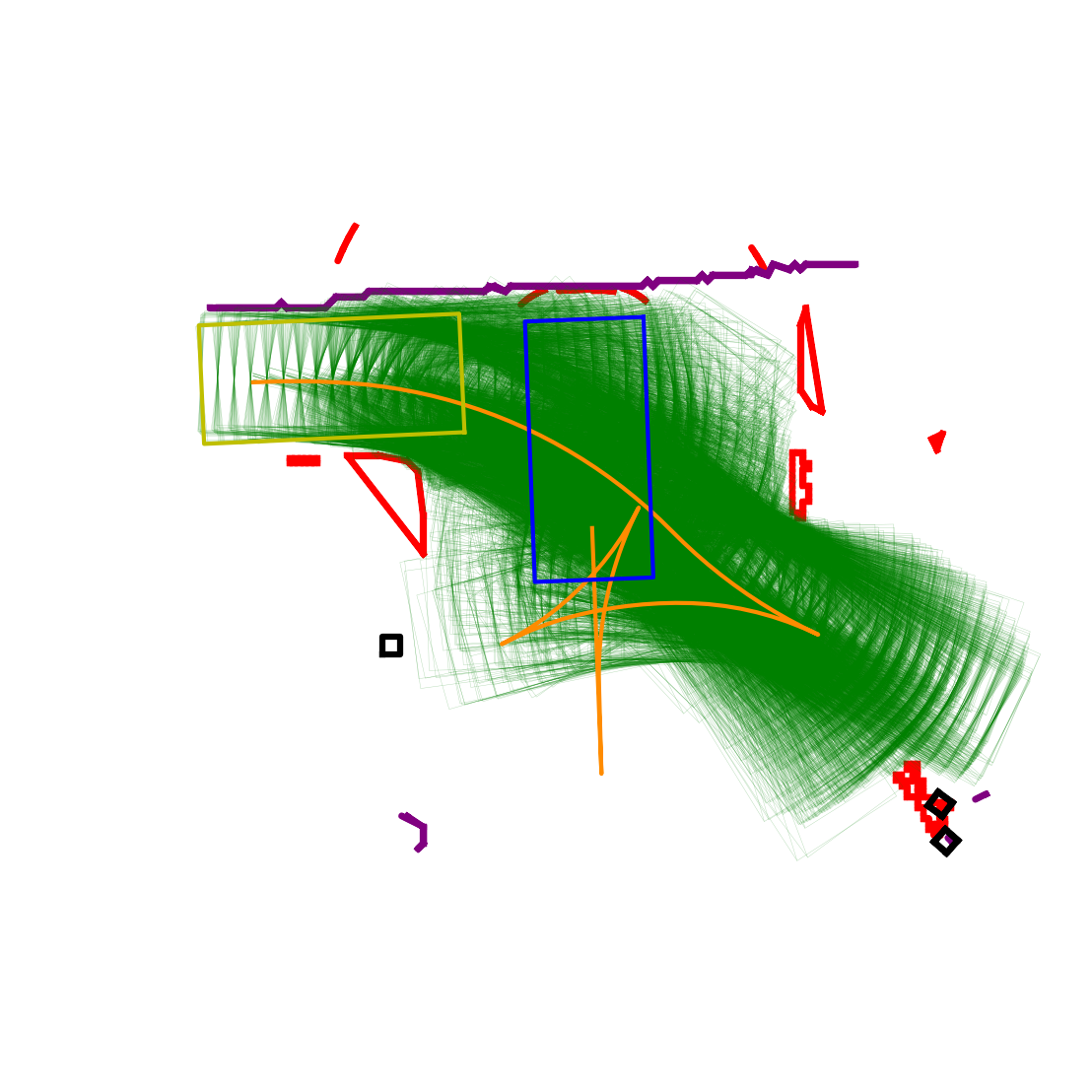}}};

         \draw (2*\xoffset, 0) node (p21) {\fcolorbox{black}{black}{\includegraphics[width=\fwidth cm]{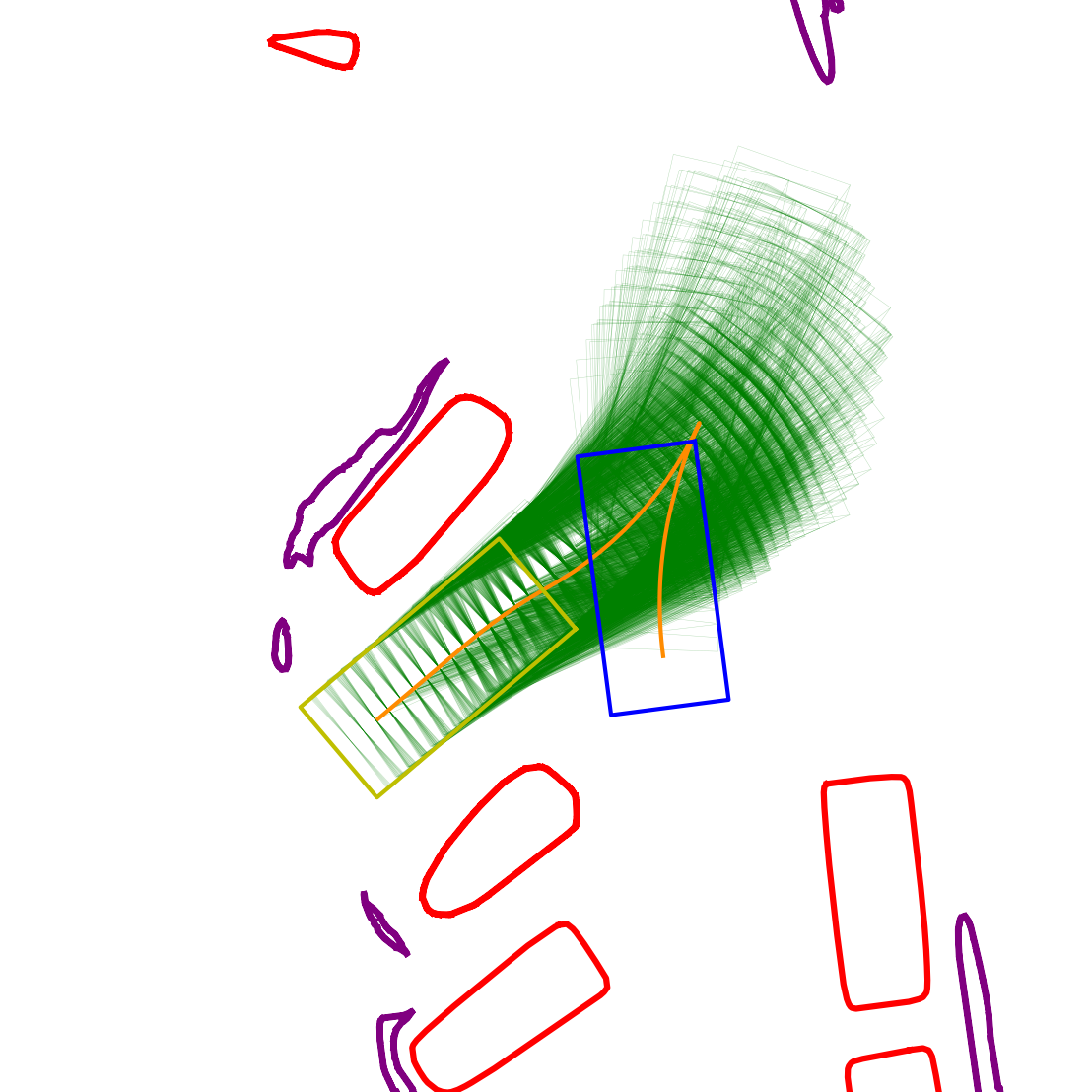}}};
         \draw (2*\xoffset, -\yoffset) node (p22) {\fcolorbox{black}{black}{\includegraphics[width=\fwidth cm]{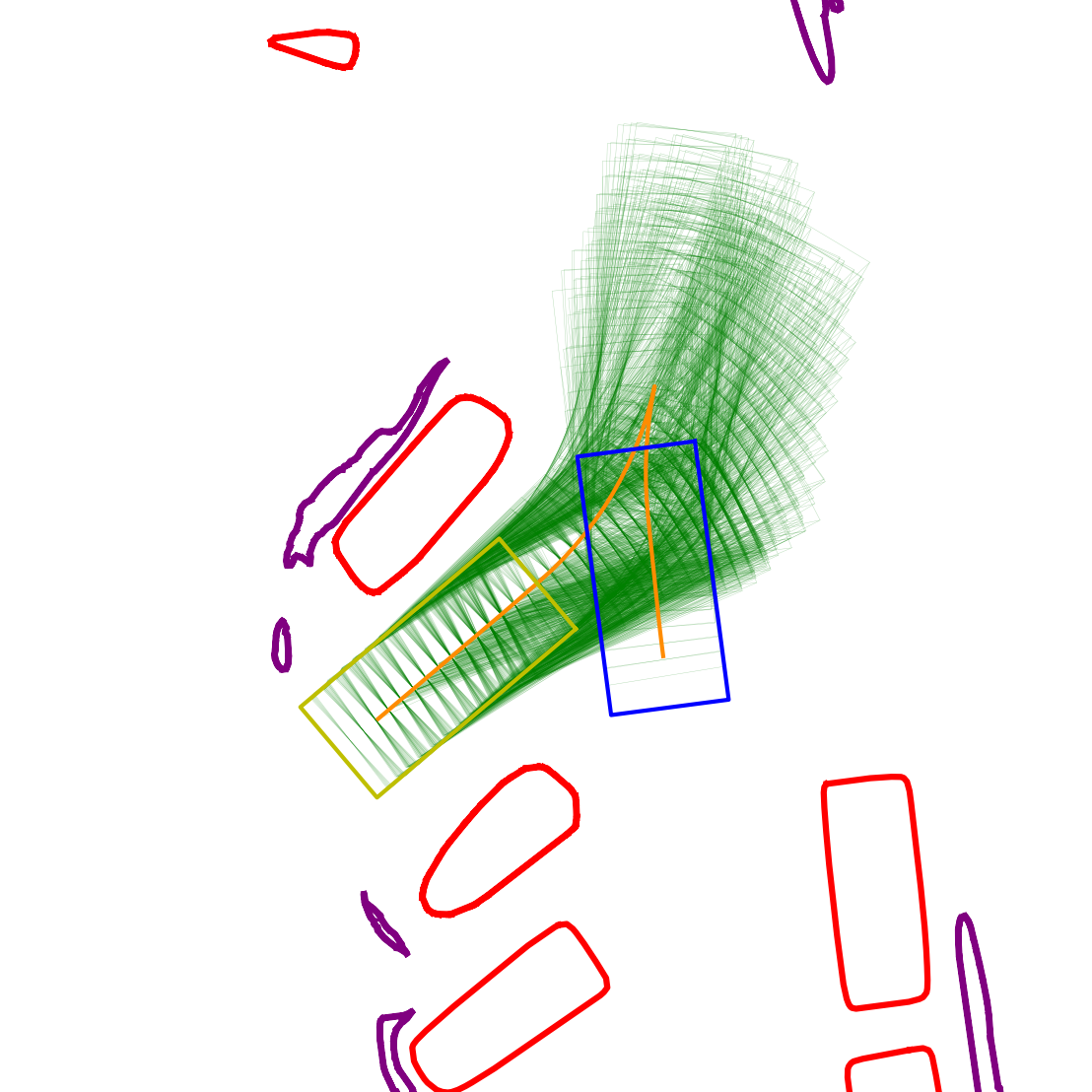}}};

         \node[] at (0, -1.6*\yoffset) {\scriptsize (a) Parallel Parking};
         \node[] at (\xoffset, -1.6*\yoffset) {\scriptsize (b) Perpendicular Parking};
         \node[] at (2*\xoffset, -1.6*\yoffset) {\scriptsize (c) Diagonal Parking};

      \end{tikzpicture}
   \caption{Comparison between Hybrid A* (above) and MCTS (below) results. The red, purple, and black polygons are different types of obstacles. The agent needs to park from the blue box to the yellow box. The orange curves are the generated path while the pale green boxes are the visited search nodes respectively.}
   \label{figurelabel6}
   \end{figure}
\end{center}

\subsection[short]{Parallel parking with narrow spot in length}

As shown in Fig. 6 (a), Our agent would like to take a parallel parking into a spot that is close to a curb and narrow in length. 
The Hybrid A* makes a huge effort to manage to get out from the spot, leading to unnecessary gear changes within the spot.
The MCTS however, uses its prior knowledge to navigate out quickly with clean moves, nudging the surrounding obstacles and makes to the 
destination confidently with no other tryouts.

\subsection[short]{Perpendicular parking under complex environment}

Fig. 6 (b) records a rare situation where our agent would like to park vertically into a spot, while there is a dead end ahead and numerous obstacles around. Under such a scenario, no well-rounded heuristic function could be applied 
to Hybrid A*, reducing its exploitation preference nearly to the level of Dijkstra's. The search space gets almost to exhaustion making the planning time very long. MCTS well balances its exploitation and exploration tendency in this 
case, managed to generate a human-like path with less than 500ms.

\subsection{Diagonal parking with enough space}
When it comes to an easier task where a clear exploitation suggestion like the one in Fig. 6 (c), MCTS still has a solid performance compared to the result of Hybrid A*, 
generating similar solutions within the same amount of time.

\section{CONCLUSION}

In this paper, we present a method that integrates reinforcement learning into the Monte Carlo tree search algorithm to expedite automated parking tasks. We formulate path planning as a Markov Decision Process 
with a bicycle model as the vehicle kinematics. Through the design of the Monte Carlo search tree structure and relative strategies in its iterative steps, we ensure adaptability within the path planning scheme. Furthermore,  a neural network is used to 
generate the policy distribution and estimate the corresponding value of a state within the MCTS. We implement a reinforcement learning pipeline with MCTS serving as a policy improvement operator on our generated dataset, and achieved an excellent training 
result. The algorithm has been successfully deployed on real-time autonomous vehicles and greatly enhances the experience of the automated parking product.

\begingroup
\footnotesize
\bibliographystyle{IEEEtran}
\bibliography{references}
\endgroup

\end{document}